\documentclass[pdflatex,sn-basic]{sn-jnl}% Basic Springer Nature Reference Style/Chemistry Reference Style

\usepackage{graphicx}%
\usepackage{multirow}%
\usepackage{amsmath,amssymb,amsfonts}%
\usepackage{amsthm}%
\usepackage{mathrsfs}%
\usepackage[title]{appendix}%
\usepackage{textcomp}%
\usepackage{manyfoot}%
\usepackage{booktabs}%
\usepackage{algorithm}%
\usepackage{algorithmicx}%
\usepackage{algpseudocode}%
\usepackage{listings}%
\usepackage[utf8]{inputenc} % allow utf-8 input
\usepackage[T1]{fontenc}    % use 8-bit T1 fonts
\usepackage{hyperref}       % hyperlinks
\usepackage{url}            % simple URL typesetting
\usepackage{nicefrac}       % compact symbols for 1/2, etc.
\usepackage{microtype}      % microtypography

\usepackage{caption}
\usepackage{subcaption}
\usepackage{longtable}

\usepackage{mathtools}
\usepackage{setspace}
\usepackage{multicol}
\usepackage[dvipsnames]{xcolor}

\usepackage{listings}
\usepackage{afterpage}

\newcommand{\red}[1]{\textcolor{black}{#1}}

\theoremstyle{thmstyleone}%
\theoremstyle{thmstyletwo}%

\theoremstyle{thmstylethree}%

\begin{document}

\title[Article Title]{XAI-TRIS: Non-linear \red{image} benchmarks to quantify \red{false positive post-hoc attribution of feature importance}}


\author*[1]{\fnm{Benedict} \sur{Clark}}\email{benedict.clark@ptb.de}

\author*[2]{\fnm{Rick} \sur{Wilming}}\email{rick.wilming@tu-berlin.de}

\author*[1,2]{\fnm{Stefan} \sur{Haufe}}\email{haufe@tu-berlin.de}

\affil[1]{\orgname{Physikalisch-Technische Bundesanstalt}, \orgaddress{\street{Abbestr. 2-12}, \postcode{10587}, \city{Berlin}, \country{Germany}}}

\affil[2]{\orgname{Technische Universität Berlin}, \orgaddress{\street{Str. des 17. Juni 135}, \postcode{10623}, \city{Berlin}, \country{Germany}}}

% \affil[3]{\orgdiv{Department}, \orgname{Organization}, \orgaddress{\street{Street}, \city{City}, \postcode{610101}, \state{State}, \country{Country}}}

% \author{%
%  Benedict Clark \\
%   Physikalisch-Technische Bundesanstalt\\
%   Abbestr. 2–12 10587 Berlin, Germany \\
%   \texttt{benedict.clark@ptb.de} \\
%   \And
%   Rick Wilming, Stefan Haufe \\
%   Technische Universität Berlin \\
%    Str. des 17. Juni 135, 10623 Berlin, Germany \\
%    \texttt{\{rick.wilming, haufe\}@tu-berlin.de}
%   % \And
%   % Stefan Haufe \\
%   % Technische Universität Berlin \\
%   %  Str. des 17. Juni 135, 10623 Berlin, Germany \\
%   % \texttt{haufe@tu-berlin.de} \\
% }


\abstract{
The field of  `explainable' artificial intelligence (XAI) has produced highly \red{acclaimed} methods that seek to make the decisions of complex machine learning (ML) methods `understandable' to humans, for example by attributing `importance' scores to input features. 
Yet, a lack of formal underpinning leaves it unclear as to what conclusions can safely be drawn from the results of a given XAI method and has also so far hindered the theoretical verification and empirical validation of XAI methods.
This means that challenging non-linear problems, typically solved by deep neural networks, presently lack appropriate remedies.
Here, we craft benchmark datasets for \red{one linear and} three different non-linear classification scenarios, in which the important class-conditional features are known by design, serving as ground truth explanations. 
Using novel quantitative metrics, we benchmark the explanation performance of a wide set of XAI methods across three deep learning model architectures. 
We show that popular XAI methods are often unable to significantly outperform random performance baselines and edge detection methods, \red{attributing false-positive importance to features with no statistical relationship to the prediction target rather than truly important features}. 
Moreover, we demonstrate that explanations derived from different model architectures can be vastly different; thus, prone to misinterpretation even under controlled conditions. 
}




\maketitle

\section{Introduction}
Only recently, a trend towards the objective empirical validation of XAI methods using ground truth data has been observed \citep{tjoaQuantifyingExplainabilitySaliency2020, li2021experimental, zhouFeatureAttributionMethods2022,arrasCLEVRXAIBenchmark2022,gevaertEvaluatingFeatureAttribution2022,agarwalOpenXAITowardsTransparentEvaluation2022}. These studies are, however, limited in the extent to which they permit a quantitative assessment of explanation performance, in the breadth of XAI methods evaluated, and in the difficulty of the posed `explanation' problems. 
%As an example, an image based benchmark placing an object on a constant background would allow for an easy `explanation' by simply extracting the object or its outline from the background, which is possible without ML models using standard image processing. 
In particular, most published benchmark datasets are constructed in a way such that realistic correlations between class-dependent (e.g., the foreground or object of an image) and class-agnostic (e.g., the image background) features are excluded.
In practice, such dependencies can give rise to features acting as suppressor variables.
Briefly, suppressor variables have no statistical association to the prediction target on their own, yet including them may allow an ML model to remove unwanted signals (noise), which can lead to improved predictions. 
In the context of image or photography data, suppressor variables could be parts of the background that capture the general lighting conditions. A model can use such information to normalize the illumination of the object and, thereby, improve object detection. 
% In the case of histology or microscopy data, pixels can become suppressor variables if they are part of a class-unspecific structure or cell that extends into and overlays or occludes another, class-specific, structure at a different location. 
More details on the principles of suppressor variables can be found in \citet{congerRevisedDefinitionSuppressor1974,friedmanGraphicalViewsSuppression2005,haufeInterpretationWeightVectors2014,wilmingScrutinizingXAIUsing2022}.
Here we adopt the formal requirement that an input feature should only be considered important if it has a statistical association with the prediction target, or is associated to it by construction. In that sense, it is undesirable to attribute importance to pure suppressor features.

Yet, \citet{wilmingScrutinizingXAIUsing2022} have shown that some of the most popular model-agnostic XAI methods are susceptible to the influence of suppressor variables, even in a linear setting. 
Using synthetic linearly separable data defining an explicit ground truth for XAI methods and linear models, \citet{wilmingScrutinizingXAIUsing2022} showed that a significant amount of feature importance is incorrectly attributed to suppressor variables. 
They proposed quantitative performance metrics for an objective validation of XAI methods, but limited their study to linearly separable problems and linear models. 
They demonstrated that methods based on so-called activation patterns (that is, univariate mappings from predictions to input features), based on the work of \citet{haufeInterpretationWeightVectors2014}, provide the best explanations. 
\red{\citet{wilmingTheoreticalBehavior2023} took this one step further and presented a minimal two-dimensional linear example, analytically showing that many popular XAI methods attribute arbitrarily high importance to suppressor variables.} 
However, it is unclear as to what extent these results would transfer to various non-linear settings. 

Thus, well-designed non-linear ground truth data comprising of realistic correlations between important and unimportant features are needed to study the influence of suppressor variables on XAI explanations in non-trivial settings, which is the purpose of this paper. We go beyond existing work in the following ways:  

\textbf{First}, we design one linear and three non-linear binary image classification problems, in which different types and combinations of tetrominoes \citep{golombPolyominoesPuzzles1996}, overlaid on a noisy background, need to be distinguished. In all cases, ground truth explanations are explicitly known through the location of the tetrominoes. 
Apart from the linear case, these classification problems require (different types of) non-linear predictive models to be solved effectively. 
    
\textbf{Second}, based on signal detection theory and optimal transport, we define \red{three} suitable quantitative metrics of `explanation performance' designed to handle the case of few important features.
    %while prioritizing false-positive over false negative detections of important features.
    
\textbf{Third}, using three different types of background noise (white, correlated, imagenet), we invoke the presence of suppressor variables in a controlled manner and study their effect on explanation performance. 

\textbf{Fourth}, we evaluate the explanation performance of no less than sixteen of the most popular model-agnostic and model-specific XAI methods, across three different machine learning architectures.

Finally, we propose four model-agnostic baselines that can serve as null models for explanation performance.
\section{Methods}\label{sec:methods}
% Our experimental pipeline starts with the creation of synthetic datasets for four binary image classification problems across two types of background noise, such that we have eight potential scenarios. These data are used to train three types of classifiers which each present different characteristics and differing abilities to appropriately model each scenario. Trained models are then subjected to a plethora of XAI methods to obtain feature `importance' (a.k.a. `saliency') maps for test data, which are then evaluated by two novel quantitative performance metrics. 
%
\subsection{Data generation}
For each scenario, we construct an individual dataset of $64 \times 64$-sized images as $\mathcal{D} = {(\mathbf{x}^{(n)},y^{(n)})}_{n=1}^N$, consisting of \textit{i.i.d} observations $(\mathbf{x}^{(n)} \in \mathbb{R}^D, y^{(n)} \in \{0,1\})_{n=1}^N$, where feature space $D=64^2=4096$ and $N=40,000$. Here, $\mathbf{x}^{(n)}$ and $y^{(n)}$ are realizations of the random variables $\mathbf{X}$ and $Y$, with joint probability density function $p_{\mathbf{X},Y}(\mathbf{x},y)$. 
% \citeauthor{wilmingScrutinizingXAIUsing2022} define a known set of important features 
% %
% \begin{equation}
%     \label{eq:set-of-important-features}
%     \mathcal{F}^+ \coloneqq \{ d \mid X_d \dep Y\},
% \end{equation}
% %
% in which input feature $\mathbf{x}_d$ statistically depends on the target variable $y$. We will use this to form our ground truth for each explanation. 

In each scenario, we generate a sample $\mathbf{x}^{(n)}$ as a combination of a signal pattern $\boldsymbol{a}^{(n)}\in \mathbb{R}^{D}$, carrying the set of truly important features used to form the ground truth for an ideal explanation, with some background noise $\boldsymbol{\eta}^{(n)} \in \mathbb{R}^{D}$. 
We follow two different generative models depending on whether the two components are combined additively or multiplicatively. 
\paragraph{Additive generation process}
For additive scenarios, we define the data generation process
\begin{equation}
    \label{eq:additive-data-generation}
    \mathbf{x}^{(n)} = \alpha (R^{(n)} \circ (H \circ \boldsymbol{a}^{(n)})) + (1-\alpha) (G \circ \boldsymbol{\eta}^{(n)}), 
\end{equation}
for the $n$-th sample. Signal pattern $\red{\boldsymbol{a}}^{(n)} =\red{\boldsymbol{a}}(y^{n})$ carries differently shaped tetromino patterns depending on the binary class label $y^{(n)} \sim \text{Bernoulli(\nicefrac{1}{2})}$. We apply a 2D Gaussian spatial smoothing filter $H:\mathbb{R}^D \to \mathbb{R}^D$ to the signal component to smooth the integration of the pattern's edges into the background, with smoothing parameter (spatial standard deviation of the Gaussian) $\sigma_{\text{smooth}}=1.5$. 
The Gaussian filter $H$ can technically provide infinite support to $\boldsymbol{a}^{(n)}$, so in practice we threshold the support at $5\%$ of the maximum level. 
White Gaussian noise $\boldsymbol{\eta}^{(n)} \sim \mathcal{N}(\mathbf{0}, \mathbf{I}_D)$, representing a non-informative background, is sampled from a multivariate normal distribution with zero mean and identity covariance $\mathbf{I}_D$. 
% This results in independent standard-normal distributed noise for each feature dimension. This background scenario is denoted as WHITE. 
For each classification problem, we define a second background scenario, denoted as CORR, in which we apply a separate 2D Gaussian spatial smoothing filter $G:\mathbb{R}^D \to \mathbb{R}^D$ to the noise component $\boldsymbol{\eta}^{(n)}$. 
Here, we set the smoothing parameter to $\sigma_{\text{smooth}}=10$. 
The third background type is that of samples from the ImageNet database \citep{deng2009imagenet}, denoted IMAGENET. We scale and crop images to be $64 \times 64$-px in size, preserving the original aspect ratio. Each 3-channel RGB image is converted to a single-channel gray-scale image using the built-in Python Imaging Library (PIL) functions and is zero-centered by subtraction of the sample's mean value.

As alluded to below, we also analyze a scenario where the signal pattern $\boldsymbol{a}^{(n)}$ underlies a random spatial rigid body (translation and rotation) transformation $R^{(n)}: \mathbb{R}^D \to \mathbb{R}^D$. 
All other scenarios make use of the identity transformation $R^{(n)} \circ (H \circ \boldsymbol{a}^{(n)}) = H \circ \boldsymbol{a}^{(n)}$.
Transformed signal and noise components $(R^{(n)} \circ (H \circ \boldsymbol{a}^{(n)}))$ and $(G \circ \boldsymbol{\eta}^{(n)})$ are horizontally concatenated into matrices $\mathbf{A} = \left[ (R^{(1)} \circ (H \circ \boldsymbol{a}^{(1)})), \hdots, (R^{(N)} \circ (H \circ \boldsymbol{a}^{(N)})) \right]$ and $\mathbf{E} = \left[ (G \circ \boldsymbol{\eta}^{(1)}), \hdots, (G \circ \boldsymbol{\eta}^{(N)}) \right]$.
Signal and background components are then normalized by the Frobenius norms of $\mathbf{A}$ and $\mathbf{E}$: $R^{(n)} \circ (H \circ \boldsymbol{a}^{(n)})) \leftarrow \nicefrac{(R^{(n)} \circ (H \circ \boldsymbol{a}^{(n)}))}{||\mathbf{A}||_{\text{F}}}$ and
$(G \circ \boldsymbol{\eta}^{(n)}) \leftarrow \nicefrac{(G \circ \boldsymbol{\eta}^{(n)})}{||\mathbf{E}||_{\text{F}}}$, where the Frobenius norm of a matrix $\mathbf{A}$ is defined as $||\mathbf{A}||_{\text{F}} \coloneqq (\sum^N_{n=1}\sum^D_{d=1}(\boldsymbol{a}^{(n)}_d)^2)^{1/2}$. 
% Transformed signal and background components are then normalized by the Frobenius norms, defined as $||\mathbf{A}||_{\text{F}} \coloneqq (\sum^N_{n=1}\sum^D_{d=1}(\boldsymbol{a}^{(n)}_d)^2)^{1/2}$ for a given matrix $\mathbf{A} \in \mathbb{R}^{N \times D}$.
Finally, a weighted sum of the signal and background components is calculated, where the scalar parameter $\alpha \in [0,1]$ determines the signal-to-noise ratio (SNR).
\paragraph{Multiplicative generation process}
For multiplicative scenarios, we define the generation process
\begin{equation}
    \label{eq:multiplicative-data-generation}
    \mathbf{x}^{(n)} = \left(\boldsymbol{1} - \alpha \left(R^{(n)} \circ  (H^{(n)} \circ \boldsymbol{a}^{(n)})) \right)\right)\left(G \circ \boldsymbol{\eta}^{(n)}\right) \;, 
\end{equation}
where $\boldsymbol{a}^{(n)}$, $\boldsymbol{\eta}^{(n)}$, $R^{(n)}$,  $H$ and $G$ are defined as above, $\mathbf{A}$ and $\mathbf{E}$ are Frobenius-normalized, and $\boldsymbol{1} \in \mathbb{R}^D$. 

For data generated via either process, we scale each sample $\mathbf{x}^{(n)} \in \mathbb{R}^D$ to the range $[-1,1]^D$, such that $\mathbf{x}^{(n)} \leftarrow \mathbf{x}^{(n)} / \max|\mathbf{x}|$, where $\max|\mathbf{x}|$ is the maximum absolute value of \red{any feature across the dataset}. 
\paragraph{Emergence of suppressors}
Note that the correlated background noise scenario induces the presence of suppressor variables, both in the additive and the multiplicative data generation processes. 
A suppressor here would be a pixel that is not part of the foreground $R^{(n)} \circ (H \circ \boldsymbol{a}^{(n)})$, but whose activity is correlated with a pixel of the foreground by virtue of the smoothing operator $G$. 
Based on previously reported characteristics of suppressor variables \citep{congerRevisedDefinitionSuppressor1974,friedmanGraphicalViewsSuppression2005,haufeInterpretationWeightVectors2014,wilmingScrutinizingXAIUsing2022}, we expect that XAI methods may be prone to attributing importance to suppressor features in the considered linear and non-linear settings, leading to drops in explanation performance as compared to the white noise background setting.
\paragraph{Scenarios}
We make use of tetrominoes \citep{golombPolyominoesPuzzles1996}, geometric shapes consisting of four blocks (each block here being $8 \times 8$-pixels), to define each signal pattern $\boldsymbol{a}^{(n)} \in \mathbb{R}^{64 \times 64}$. 
We choose these as the basis for signal patterns as they allow a fixed and controllable amount of features (pixels) per sample, and specifically the `T'-shaped and `L' shaped tetrominoes due to their four unique appearances under each 90-degree rotation.
These induce statistical associations between features and target in four different binary classification problems:
%
% We use a `T'-shaped tetromino near the top left corner if $y=0$ and an `L'-shaped tetromino near the bottom-right corner if $y=1$, leading to a binary classification problem. 
% The signal pattern $\boldsymbol{a}$ is identical across all scenarios.
% The signal pattern induces statistical associations between features and target in the generated synthetic datasets, and, following \citeauthor{wilmingScrutinizingXAIUsing2022}, we use those to determine the set of important features to form our ground truth 
%
%
\paragraph{Linear (LIN) and multiplicative (MULT)} 
For the linear case, we use the additive generation model Eq.~\eqref{eq:additive-data-generation}, and for the multiplicative case, we instead use the multiplicative generation model.
In both, signal patterns are defined as a `T'-shaped tetromino pattern $\red{\boldsymbol{a}}^{\text{T}}$ near the top left corner if $y=0$ and an `L'-shaped tetromino pattern $\red{\boldsymbol{a}}^{\text{L}}$ near the bottom-right corner if $y=1$, leading to the binary classification problem. 
Each pattern is encoded such that $a^{\text{T/L}}_{i,j} = 1$ for each pixel in the tetromino pattern, positioned at the $i$-th row and $j$-th column of $\boldsymbol{a}^{\text{T/L}}$, and zero otherwise.
\paragraph{Translations and rotations (RIGID)} 
In this scenario, $a^{\text{T/L}}$ defining each class are no longer in fixed positions but are randomly translated and rotated by multiples of 90 degrees according to a rigid body transform $R^{(n)}$, constrained such that the entire tetromino is contained within the image. In contrast to the other scenarios, we use a 4-pixel thick tetromino here to enable a larger set of transformations, and thus increase the complexity of the problem.
This is an additive manipulation in accordance with (\ref{eq:additive-data-generation}).
\paragraph{XOR} 
The final scenario is that of an additive XOR problem, where we use both tetromino variants $a^{\text{T/L}}$ in every sample. Transformation $R^{(n)}$ is, once again, the identity transform here.
Class membership is defined such that members of the first class, where $y=0$, combine both tetrominoes with the background of the image either positively or negatively, such that $\red{\boldsymbol{a}}^{\text{XOR++}} = \red{\boldsymbol{a}}^{\text{T}} + \red{\boldsymbol{a}}^{\text{L}}$ and $\red{\boldsymbol{a}}^{\text{XOR-{}-}} = - \red{\boldsymbol{a}}^{\text{T}} - \red{\boldsymbol{a}}^{\text{L}}$.
Members of the opposing class, where $y=1$, imprint one shape positively, and the other negatively, such that $\red{\boldsymbol{a}}^{\text{XOR+-}} = \red{\boldsymbol{a}}^{\text{T}} - \red{\boldsymbol{a}}^{\text{L}}$ and $\red{\boldsymbol{a}}^{\text{XOR-+}} = - \red{\boldsymbol{a}}^{\text{T}} + \red{\boldsymbol{a}}^{\text{L}}$. 
Each of the four XOR cases are equally frequently represented across the dataset. 

\begin{figure}[!ht]
  \centering
    \includegraphics[width=\textwidth]{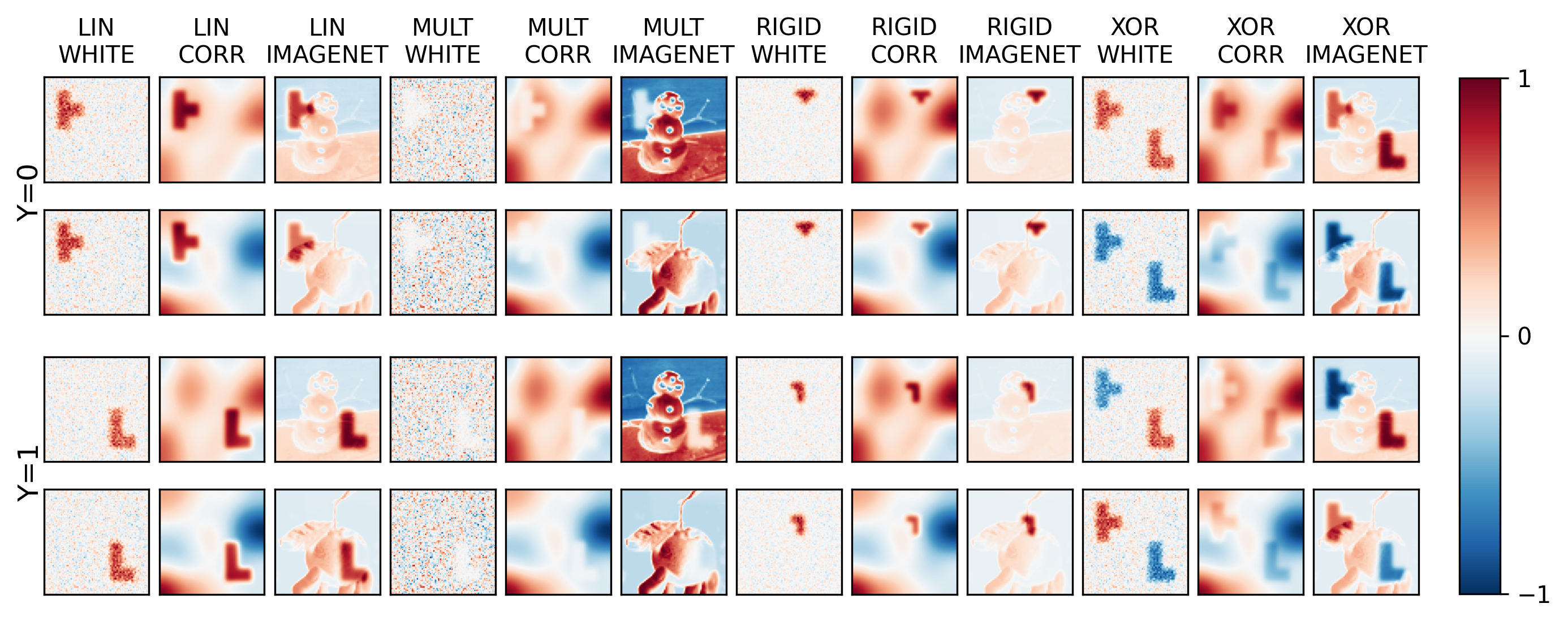}
    \caption{Examples of data for each scenario, showing differences between samples of each class.}
    \label{fig:data_plot}
  % \vskip -0.1in
\end{figure}

Figure~\ref{fig:data_plot} shows two examples from each class of each classification problem and for the three background types -- Gaussian white noise (WHITE), smoothed Gaussian white noise (CORR), and ImageNet samples (IMAGENET). 
Figure~\ref{fig:snr_plot} in the supplementary material shows examples of each of the 12 scenarios across four signal-to-noise ratios (SNRs). 
% As SNR increases, saliency of the signal pattern very quickly becomes obvious. 
%

With each classification scenario defined, we can form the ground truth feature set of important pixels for a given input based on the positions of tetromino pixels as
\begin{equation}
    \label{eq:ground truth}
    \mathcal{F}^+(\mathbf{x}^{(n)}) \coloneqq \left\{d \mid \left( R^{(n)} \circ (H \circ \boldsymbol{a}^{(n)}) \right)_d \neq 0, \, d \in \left\{1, \dots, 4096\right\} \right\} \;.
\end{equation}
For the LIN and MULT scenarios, each sample either contains a `T' or an `L' tetromino at a fixed position, corresponding to the fixed patterns $\red{\boldsymbol{a}}^{\text{T}}$ and $\red{\boldsymbol{a}}^{\text{L}}$. Since the absence of a tetromino at one location is just as informative as the presence of the other at another location, we augment the set of important pixels for these two settings as 
\begin{equation}
    \label{eq:ground truth_fixed8}
    \mathcal{F}^+(\mathbf{x}^{(n)}) \coloneqq \left\{d \mid H \circ \boldsymbol{a}^{\text{T}}_d \neq 0 \lor H \circ \boldsymbol{a}^{\text{L}}_d \neq 0, \, d \in \{1, \dots, 4096\} \right\} \;.
\end{equation}
Note that this definition is equivalent to Eq.~\eqref{eq:ground truth} for the XOR scenario. Moreover, it is equivalent to an operationalization of feature importance put forward by \citet{wilmingScrutinizingXAIUsing2022} for the three static scenarios LIN, MULT, and XOR. \citet{wilmingScrutinizingXAIUsing2022} define any feature as important if it has a statistical dependency to the prediction target across the studied sample. In all cases, an ideal explanation method should attribute importance only to members of the set $\mathcal{F}^+(\mathbf{x}^{(n)})$.

For training each model and the subsequent analyses, we divide each dataset three-fold by a $90/5/5$ split into a training set $\mathcal{D}_{\text{train}}$, a validation set $\mathcal{D}_{\text{val}}$, and a test set $\mathcal{D}_{\text{test}}$.
% such that $N^{\text{train}} = 8000$, $N^{\text{val}} = 1000$, and $N^{\text{test}} = 1000$, resulting in $N=10000$ samples per dataset. 
%
%
\subsection{Classifiers}\label{sec:classifiers}
We use three architectures to model each classification problem. Firstly, a Linear Logistic Regression (LLR) model, which is a single-layer neural network with two output neurons and a softmax activation function. 
Secondly, a Multi-Layer Perceptron (MLP) with four fully-connected layers, where each of the hidden layers uses Rectified Linear Unit (ReLU) activations. 
The two-neuron output layer is once again softmax-activated. Finally, we define a Convolutional Neural Network (CNN) with four blocks of ReLU-activated convolutional layers followed by a max-pooling operation, with a softmax-activated two-neuron output layer. 
The convolutional layers are specified with a progressively increasing amount of filters per layer $[4,8,16,32]$, a kernel size of four, a stride of one, and zero-padding.
% This padding technique is used to improve pixel utilization across each convolution, as well as to mitigate shrinking outputs of the already relatively small images, by adding extra filler pixels (set to values of zero) around the edge of each image. 
The max-pooling layers are defined with a kernel size of two and a stride of one. 
% Some popular CNN architecture features, such as batch normalization, are unavailable here due to lack of implementation support by some XAI methods. 

We train a given classifier $f^{\boldsymbol{\theta}}:\mathbb{R}^D \rightarrow \mathcal{Y}$ over parameterization $\boldsymbol{\theta}$ and $\mathcal{D}_{\text{train}}$. 
Each network is trained over 500 epochs using the Adam optimizer without regularization, with a learning rate of $0.0005$. 
The validation dataset $\mathcal{D}_{\text{val}}$ is used at each step to get a sense of how well the model is generalizing the data. 
Validation loss is calculated at each epoch and used to judge when the classifier has reached optimal performance, by storing the model state with minimum validation loss. 
This also prevents using an overfit model. 
Finally, the test dataset $\mathcal{D}_{\text{test}}$ is used to calculate the resulting model performance, and is used in the evaluation of XAI methods.
We consider a classifier to have generalized the given classification problem when the resulting test accuracy is at or above a threshold of $80\%$.

Each network is implemented in PyTorch, and also in Keras with a TensorFlow backend, so to experiment over a wider variety of XAI methods implemented using either the Captum \citep{kokhlikyanCaptumUnifiedGeneric2020} or iNNvestigate \citep{alberINNvestigateNeuralNetworks2018} frameworks. The main text focuses on the former.
%While each framework's implementation should theoretically produce equivalent discriminant functions, small differences in performance are noted, which are mitigated through the use of fixed random seeds wherever possible, as well as other experimental details outlined in Section \ref{sec:experiments}.

\subsection{XAI methods and performance baselines}
We compare sixteen popular XAI methods in our analysis. The main text focuses on the results of four: Local Interpretable Model Explanations (LIME) \citep{ribeiroWhyShouldTrust2016}, Layer-wise Relevance Propagation (LRP) \citep{bachPixelWiseExplanationsNonLinear2015}, SHapley Additive exPlanations (SHAP) \citep{lundbergUnifiedApproachInterpreting2017} and Integrated Gradients \citep{sundararajanAxiomaticAttributionDeep2017}.
% We highlight the former three methods due to their widespread adoption, whereas the latter is included due to its alleged ability to suppress the attribution of importance to suppressor variables. 

The full list is detailed in Appendix \ref{app:methods-training}. This briefly summarizes each method, and provides the details of which library was used for implementation, Captum \citep{kokhlikyanCaptumUnifiedGeneric2020} or iNNvestigate \citep{alberINNvestigateNeuralNetworks2018}, as well as the specific parameterization for each method. Generally, we follow the default parameterization for each method. 
Where necessary, we specify the baseline $\mathbf{b}$ as the zero input $\mathbf{b} = \boldsymbol{0}$, a common choice in the field \citep{mamalakisCarefullyChooseBaseline2022}.

The input to an XAI method is a model $f^{\boldsymbol{\theta}}:\mathbb{R}^D \rightarrow \mathbb{R}$, trained according to parameterization $\boldsymbol{\theta}$ over $\mathcal{D}_{\text{train}}$, the $n$-th test sample to be explained $\mathbf{x}_{\text{test}}^{(n)}$, as well as the baseline reference point $\mathbf{b} = \boldsymbol{0}$ for relevant methods. 
The method produces an `explanation' $\mathbf{s}(f^{\boldsymbol{\theta}}, \mathbf{x}_{\text{test}}^{(n)}, \mathbf{b}) \in \mathbb{R}^D$.
%
% \subsection*{Baseline methods}

We include four model-ignorant methods to generate `baseline' importance maps for comparison with the aforementioned XAI methods. 
% With this, we can judge whether these often-complicated XAI methods actually produce improved explanations over methods with no model information. 
Firstly, we consider the Sobel filter, which uses both a horizontal and a vertical filter kernel to approximate first-order derivatives of data. 
Secondly, we use the Laplace filter, which uses a single symmetrical kernel to approximate second-order derivatives of data. Both are edge detection operators, and are given for each test sample as an input. 
Thirdly, we use a sample from a random uniform distribution $U((-1,1)^D)$. 
Finally, we use the rectified test data sample $\mathbf{x}_{\text{test}}^{(n)}$ itself as an importance map.
\subsection{Explanation performance metrics}
% Based on the well-defined ground truth set of class-dependent features for a given sample $\mathcal{F}^+(\mathbf{x}^{(n)})$, we can readily form quantitative metrics that compare that set to an explanation $\mathbf{s}(f^{\boldsymbol{\theta}}, \mathbf{x}^{(n)}, \mathbf{b})$ produced by an XAI method for that sample. We define
% \begin{equation}
%     \mathbf{h}_d^{\text{true}} = \begin{cases}
%                             1, &d \in \mathcal{F}^+(\mathbf{x}^{(n)}) \\
%                             0, &d \in \mathcal{F}^-(\mathbf{x}^{(n)})
%                         \end{cases},
% \end{equation}
% where $\mathcal{F}^-(\mathbf{x}^{(n)}) \coloneqq \{1, \hdots, 64\} \setminus \mathcal{F}^+(\mathbf{x}^{(n)})$ is the set of unimportant features containing only non-class-specific background fluctuations. We define two metrics to compare $\mathbf{h}^{\text{true}}$ to each explanation $\mathbf{s}(f^{\boldsymbol{\theta}}, \mathbf{x}_{\text{test}}^{(n)}, \mathbf{b})$ produced by an XAI method for test sample $\mathbf{x}_{\text{test}}^{(n)}$, as objective and data-driven measures of explanation performance.
%
Based on the well-defined ground truth set of class-dependent features for a given sample $\mathcal{F}^+(\mathbf{x}^{(n)})$, we can readily form quantitative metrics to evaluate the quality of an explanation. 
% We define two metrics to compare $\mathcal{F}^+(\mathbf{x}_{\text{test}}^{(n)})$ to each explanation $\mathbf{s}(f^{\boldsymbol{\theta}}, \mathbf{x}_{\text{test}}^{(n)}, \mathbf{b})$, produced by an XAI method for test sample $\mathbf{x}_{\text{test}}^{(n)}$, as objective and data-driven measures of explanation performance.
%
%
\subsection*{Precision}
Omitting the sample-dependence in the notation, we define precision as the fraction of the $k=|\mathcal{F}^+|$ features of $\mathbf{s}$ with the highest absolute-valued importance scores contained within the set $\mathcal{F}^+$ itself, over the total number of important features $|\mathcal{F}^+|$ in the sample.
\red{We constrain these results to the submitted appendices, and focus on the results and analyses for the next two defined metrics.}
%
% define precision as
% $\mathrm{PREC}(\mathbf{s}, \mathbf{h}^{\text{true}}) = \mathrm{TP}/(\mathrm{TP} + \mathrm{FP})$, 
% where a true positive (TP) is given as an element of $\mathbf{s}$ contained within ground truth $\mathbf{h}^{\text{true}})$.
% is the fraction of features that are estimated to be the important by an XAI method (both true and false positives, $\mathrm{TP} + \mathrm{FP}$) that are contained in the ground truth ($\mathrm{TP}$).
%
% the fraction of truly important features amongst all features estimated to be important by an XAI method's feature importance scores
\subsection*{Earth mover's distance (EMD)}
The Earth mover's distance (EMD), also known as the Wasserstein metric, measures the optimal cost required to transform one distribution to another. We can apply this to the cost required to transform a continuous-valued importance map $\mathbf{s}$ into $\mathcal{F}^+$, where both are normalized to have the same mass. The Euclidean distance between pixels is used as the ground metric for calculating the EMD, with $\mathrm{OT}(\mathbf{s}, \mathcal{F}^+)$ denoting the cost of the optimal transport from explanation $\mathbf{s}$ to ground truth $\mathcal{F}^+$. This follows the algorithm proposed by \citet{bonneel2011displacement} and the implementation of the Python Optimal Transport library \citep{flamary2021pot}. We define a normalized EMD performance score as
\begin{equation}
    \mathrm{EMD} = 1 - \frac{\mathrm{OT}(\mathbf{s}, \mathcal{F}^+)}{\delta_{max}},
\end{equation}
where $\delta_{max}$ is the maximum Euclidean distance between any two pixels.
\paragraph*{Remark.}
Note that the ground truth $\mathcal{F}^+(\mathbf{x})$ defines the set of important pixels based on the data generation process. It is conceivable, though, that a model uses only a subset of these for its prediction, which must be considered equally correct. \red{The above} explanation performance metrics do not fully achieve invariance in that respect. However, both are designed to de-emphasize the impact of false-negative omissions of features in the ground truth on performance, while emphasizing the impact of false-positive attributions of importance to pixels not contained in the ground truth.
\red{\subsection{Importance Mass Accuracy}}
\red{Because of this, we consider a third metric, Importance Mass Accuracy (IMA). Calculated as the sum of importance attributed to the ground truth features over the total attribution in the image, this metric is akin to `Relevance mass accuracy' as defined by \citet{arrasCLEVRXAIBenchmark2022}. We calculate
\begin{equation}
    \mathrm{IMA} = \underset{s_i \in \mathcal{F}^+}{\sum_{i=1}^{|\mathcal{F}^+|}} s_i / \sum_{i=1}^{|\mathbf{s}|} s_i.
\end{equation}
This metric achieves invariance for not penalizing false negative attribution to a subset of pixels in $\mathcal{F}^+(\mathbf{x})$, whilst also utilizing the whole attribution instead of a `top-k' metric such as Precision.
Not only this, but it is a direct measure of false positive attribution, where a score of $1$ signals a perfect explanation highlighting only ground truth features as important.
We use this metric to complement the strengths of $\mathrm{EMD}$ whilst also presenting an alternative perspective to quantifying explanation performance.}
\section{Experiments}\label{sec:experiments}
Our experiments aim to answer four main questions:

\textbf{1.} Which XAI methods are best at identifying truly important features as defined by the sets $\mathcal{F}^+(\mathbf{x})$?

\textbf{2.} Does explanation performance for each method remain consistent when moving from explaining a linear classification problem to problems with different degrees of non-linearity?

\textbf{3.} Does adding correlations to the background noise, through smoothing with the Gaussian convolution filter, negatively impact explanation performance?

\textbf{4.} How does the choice of model architecture impact explanation performance?

%
% Each classification problem has three variants per problem -- one using the Gaussian random white noise background solely (WHITE), where the filter $G$ is just an identity mapping, one with smoothed noise background (CORR), where $G$ is the 2D Gaussian spatial smoothing filter with $\sigma_{\text{smooth}} = 10$, and one where the background is a sample from the ImageNet dataset. 
% This leads to 12 overall scenarios, each presenting a ranging degree of non-linearity. 
% First, a linear problem (LIN) as a point of comparison and to accompany the Linear Logistic Regression (LLR) model. 
% Second, a multiplicative classification problem (MULT), defined through \ref{eq:multiplicative-data-generation}, which shrinks signal patterns $\boldsymbol{a}$ towards zero compared to the background. 
% Third, a rigid body transformation scenario (RIGID), where $\boldsymbol{a}^\text{T/L}$ can be randomly translated and rotated according to $R$ and combined additively with the respective background noise component. 
% Finally, an XOR problem, where two tetromino patterns within $\boldsymbol{a}^\text{T/L}$ are combined additively with the background noise component in four cases as defined in the previous section. The three non-linear scenarios present differing kinds of non-linearity, and thus challenge the studied classifiers and XAI methods in different ways. 

We generate a dataset for each scenario across a range of 20 choices of $\alpha$, finding the `sweet spot' where average test accuracy over 10 trained models is at or above 80\%. 
Table \ref{tab:chosen-snrs} shows the resulting $\alpha$ values as well as the average test accuracy for each scenario, over five model trainings for datasets of size  $N=40,000$ of each scenario. 
For training each model and the subsequent analyses, we divide each dataset three-fold by an $90/5/5$ split into a training set $\mathcal{D}_{\text{train}}$, a validation set $\mathcal{D}_{\text{val}}$, and a test set $\mathcal{D}_{\text{test}}$.
From this, we compute absolute-valued importance maps $|\mathbf{s}|$ for the intersection of test data $\mathcal{D}^{\text{test}}$ correctly predicted by every appropriate classifier.
The full table of training results for finding appropriate SNRs can be seen in Appendix \ref{app:methods-training}.
\begin{table}[ht!]
\caption{Results of the model training process for each classification setting, model architecture, and background type. These results are depicted as chosen Signal-to-noise ratios (SNRs), parameterized by $\alpha$, as well as the average test accuracy (ACC, \%). }
\label{tab:chosen-snrs}
% \vskip 0.15in
\begin{small}
\begin{sc}
\begin{tabular}{lccccccc}
\toprule
& & \multicolumn{2}{c}{WHITE} & \multicolumn{2}{c}{CORR}  & \multicolumn{2}{c}{IMAGENET} \\
& & $\alpha$ & ACC & $\alpha$ & ACC & $\alpha$ & ACC \\
\midrule

        &       LLR     & $0.03$ & $89.7$  & $0.02$ & $100.0$ & $0.1$ & $87.5$ \\
\rule{0pt}{0ex}
LIN     &       MLP     & $0.03$ & $87.9$ & $0.02$ & $100.0$ & $0.1$ & $86.2$ \\
        &       CNN     & $0.03$ & $90.1$ & $0.02$ & $99.9$  & $0.1$ & $93.9$ \\
\rule{0pt}{3ex}
MULT    &    MLP        & $0.64$ & $85.8$  & $0.04$ & $89.2$  &  $0.3$   & $91.2$ \\
        &    CNN        & $0.64$ & $100.0$  & $0.04$ & $98.5$  &  $0.3$   & $91.3$ \\
\rule{0pt}{3ex}
RIGID   &    MLP        & $0.575$  & $88.9$ & $0.375$ & $99.5$ &   $0.6$   & $92.0$ \\
        &    CNN        & $0.575$ & $100.0$ & $0.375$ & $100.0$ &  $0.6$   & $99.9$ \\
\rule{0pt}{3ex}
XOR     &    MLP        & $0.1$ & $99.9$ & $0.1$ & $100.0$ &    $0.2$ & $99.9$ \\
        &    CNN        & $0.1$ & $100.0$ & $0.1$ & $100.0$  &  $0.2$ & $100.0$ \\
\bottomrule
\end{tabular}
\end{sc}
\end{small}
% \vskip -0.1in
\end{table}
Experiments were run on an internal CPU and GPU cluster, with total runtime in the order of a matter of hours.
\section{Results}\label{sec:results}
% \subsection{Qualitative analysis}
%
\begin{figure}[!ht]
    \centering
    \includegraphics[width=\textwidth]{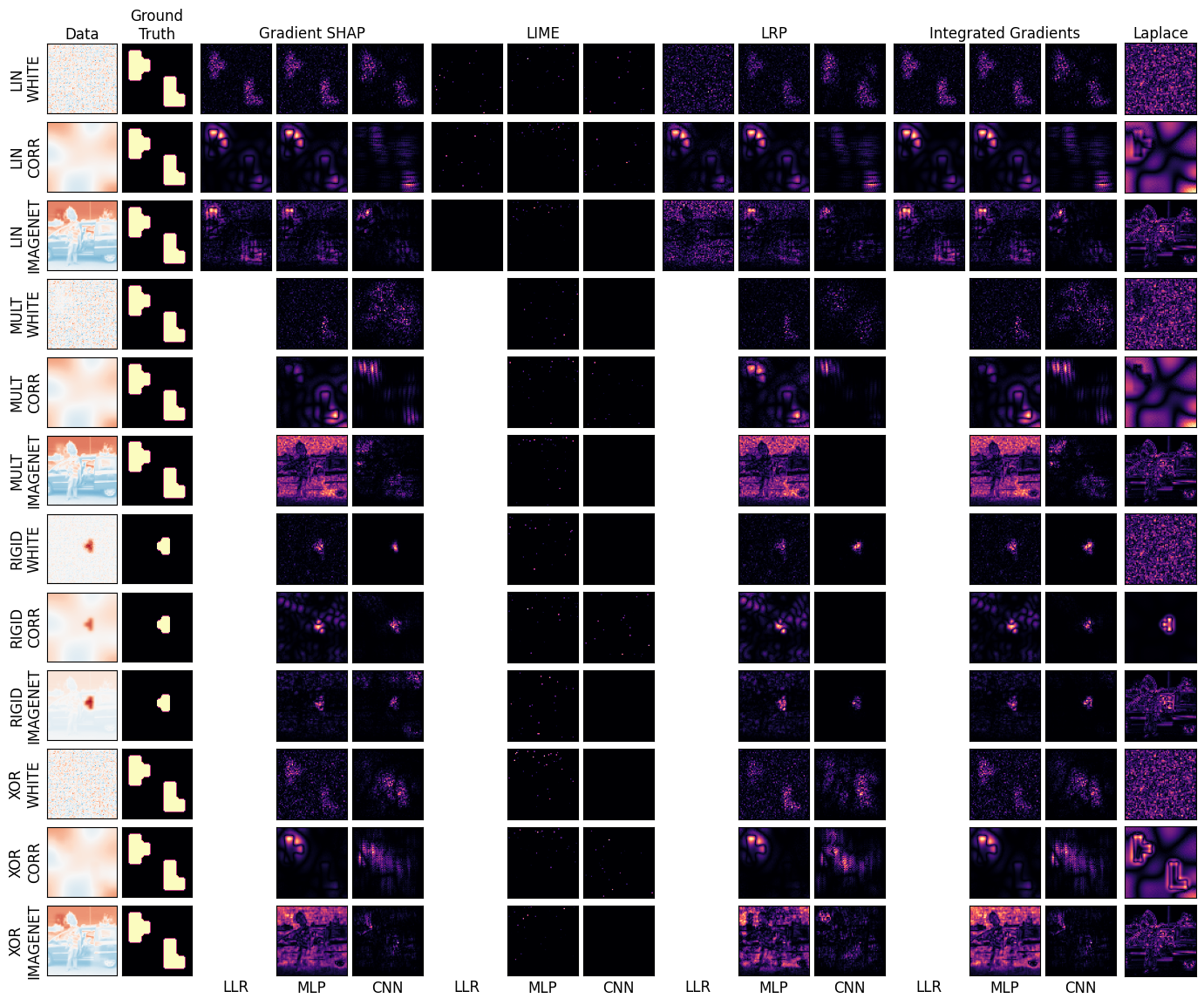}
    \caption{Absolute-valued importance maps obtained for a random correctly-predicted data sample, for selected XAI methods and baselines. Recovery of the ground truth pattern across all scenarios is best shown by XAI methods applied to a Linear Logistic Regression (LLR) model.
    The Multi-Layer Perceptron (MLP) tends to focus on noise in the case of ImageNet backgrounds, and LIME often fails to produce sensical explanations across all model architectures.
    % Note that we have also highlighted $\boldsymbol{a}^{\text{L}}$ in the ground truth for the LIN and MULT examples. We consider the opposing-class tetromino pattern $\boldsymbol{a}^{\text{L}}$ as a member of $\mathcal{F}^+$, the ground truth set of important features, because the absence of these features can be a valid reason for an XAI to highlight the respective region of the image as important.
    }
    \label{fig:qualitative-results}
    \vskip -0.2in
\end{figure}

Figure \ref{fig:qualitative-results} depicts examples of absolute-valued importance maps produced for a random correctly-predicted sample for each scenario and model. 
Shown are results for four XAI methods (Gradient SHAP, LIME, LRP, and PatternNet respectively) for each of the three models (LLR, MLP, CNN respectively) followed by the model-ignorant Laplace filter. 
Appendix \ref{app:qualitative-results} expands on the qualitative results of the main text, and Figure~\ref{fig:qualitative-results-global} shows the absolute-valued \emph{global} importance heatmaps for the LIN, MULT, and XOR scenarios, given as the mean of all explanations for every correctly-predicted sample of the given scenario and XAI method.
As the RIGID scenario has no static ground truth pattern, calculating a global importance map is not possible. 
%
% Figure \ref{fig:qualitative-results} depicts examples of saliency maps produced for a randomly chosen single data instance for each scenario and model, where the data instance was correctly predicted by the respective model. The LLR and, to some extent, the MLP appear much more likely to place importance on the opposing class as well as the target class. This is a potentially valid explanation due to the absence of the features relevant to the opposing class. Qualitatively, PatternNet and PatternAttribution appear to perform very well for the LLR and MLP, which aligns with the results shown by the authors of \cite{wilmingScrutinizingXAIUsing2022}. The explanations produced by the CNN appear more sparse and much less conclusive as to which method performs best. Despite requiring a higher SNR to model accurately, and thus a much higher saliency of the pattern within the image, the explanations produced for the Translations and Rotations scenario with the `Uncorrelated' background type are often noisier than those produced by other examples. This is especially apparent for the explanations produced for this scenario by the MLP.
%
% 
% \subsection{Quantitative analysis}

\begin{figure}[!ht]
    \centering
    \begin{subfigure}[b]{\textwidth}
       \includegraphics[width=1\linewidth]{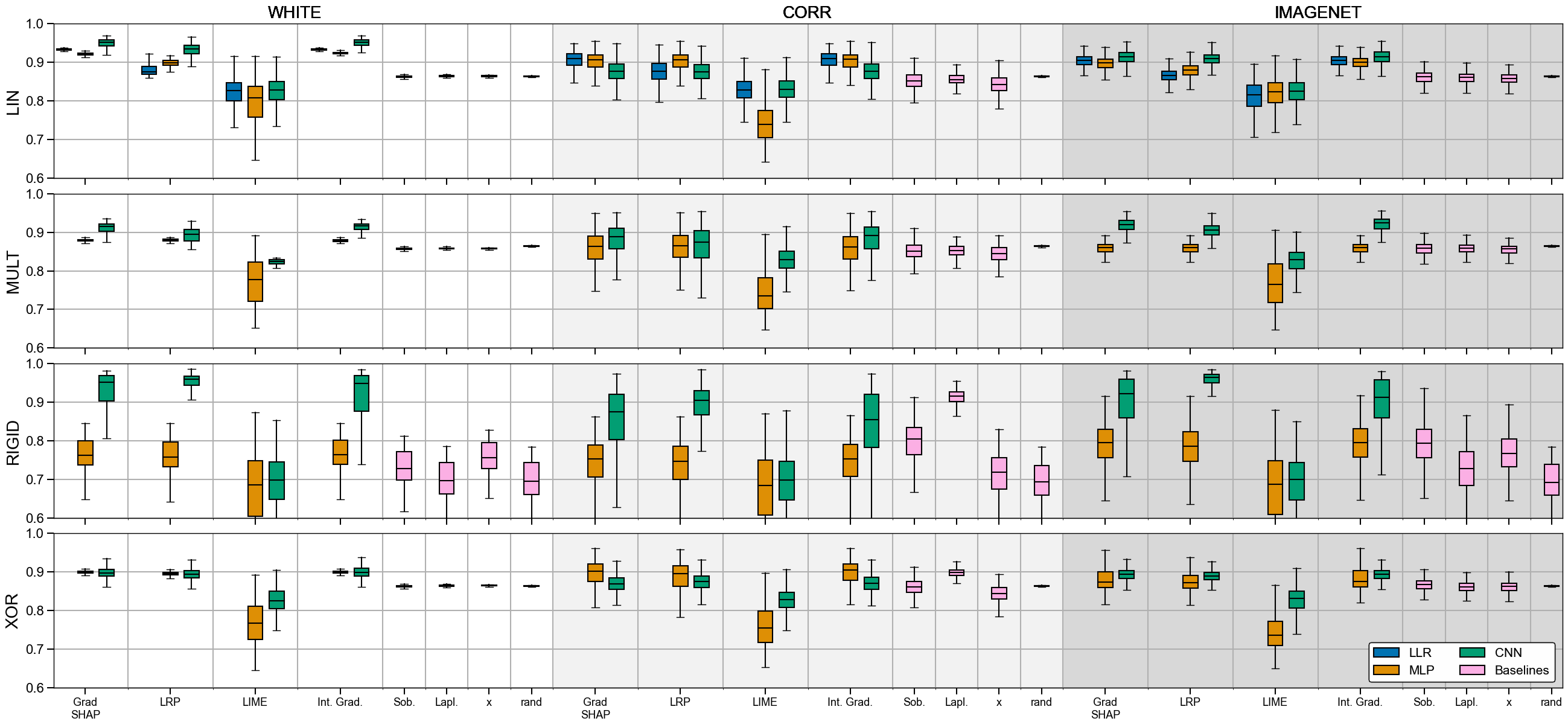}
       \caption{Earth mover's distance ($\mathrm{EMD}$)}
       \label{fig:emd-results} 
    \end{subfigure}
    
    \begin{subfigure}[b]{\textwidth}
       \includegraphics[width=1\linewidth]{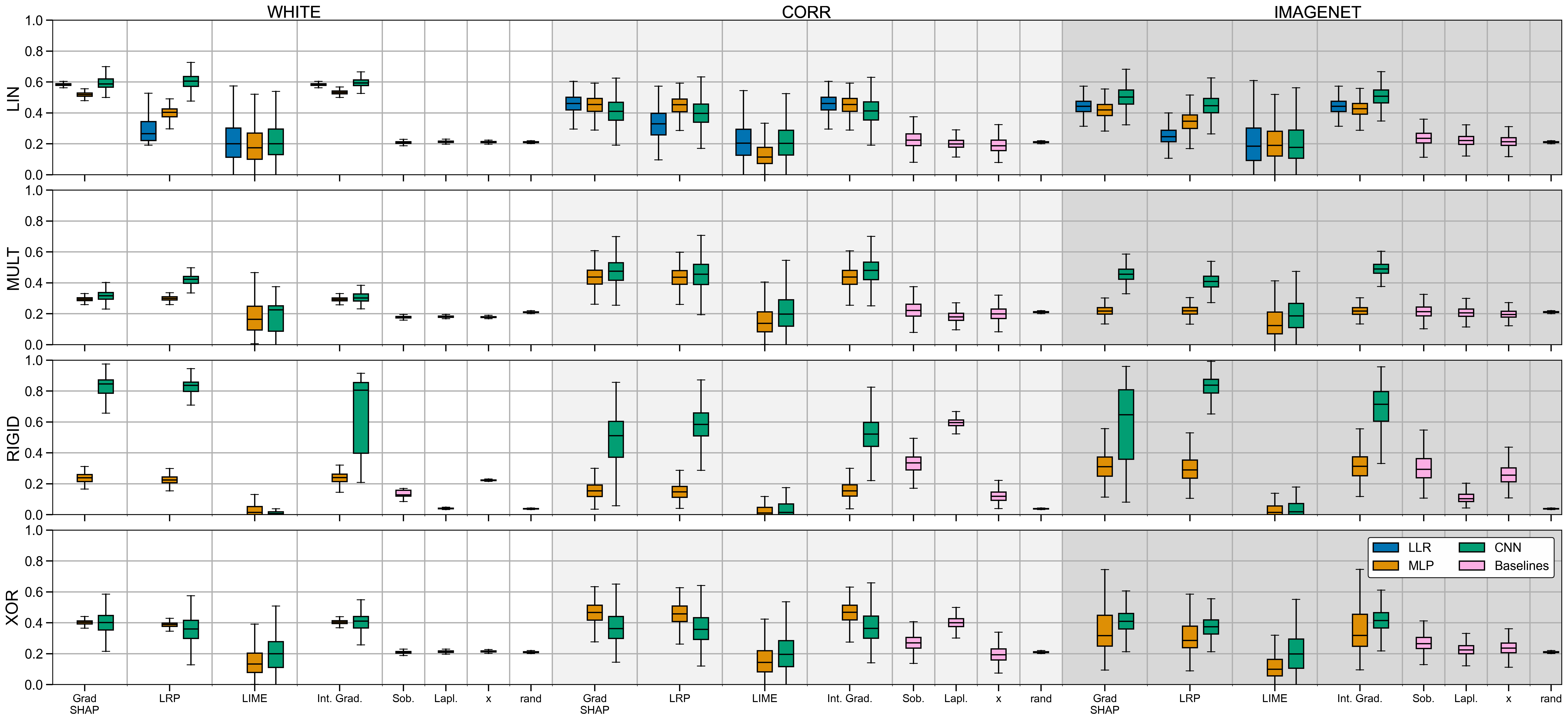}
       \caption{Importance mass accuracy ($\mathrm{IMA}$)}
       \label{fig:frac-results}
    \end{subfigure}
    \caption{Quantitative explanation performance of individual sample-based feature importance maps produced by various XAI approaches and baseline methods on correctly-predicted test samples, as per the  $\mathrm{EMD}$ \red{(top) and $\mathrm{IMA}$ (bottom)} metric\red{s}. Depicted are boxplots of median explanation performance, with upper and lower quartiles as well as outliers shown. 
    The white area\red{s} (left) show results for white background noise (WHITE), whereas the light gray shaded area\red{s} (middle) shows results for the correlated background noise (CORR) scenarios and the darker gray \red{areas} (right) for ImageNet (IMAGENET) backgrounds.}
    \label{fig:quantitative-results}
\end{figure}

% \begin{figure*}[!ht]
%     \centering
%     \includegraphics[width=\textwidth]{figures/emd_results.png}
%     \caption{Quantitative explanation performance of individual sample-based feature importance maps produced by various XAI approaches and baseline methods on correctly-predicted test samples, as per the  $\mathrm{EMD}$ metric. Depicted are boxplots of median explanation performance, with upper and lower quartiles as well as outliers shown. 
%     The white area (left) shows results for white background noise (WHITE), whereas the light gray shaded area (middle) shows results for the correlated background noise (CORR) scenarios and the darker gray (right) for ImageNet (IMAGENET) backgrounds.} 
%     \label{fig:emd-results}
%     % \vskip -0.1in
% \end{figure*}
% %
% \begin{figure*}[!h]
%     \centering
%     \includegraphics[width=\textwidth]{figures/Frac_results_hires_frac.png}
%     \caption{\red{Quantitative explanation performance of individual sample-based feature importance maps produced by various XAI approaches and baseline methods on correctly-predicted test samples, as per the $\mathrm{IMA}$ metric. Depicted are boxplots of median explanation performance, with upper and lower quartiles as well as outliers shown. 
%     The white area (left) shows results for white background noise (WHITE), whereas the light gray shaded area (middle) shows results for the correlated background noise (CORR) scenarios and the darker gray (right) for ImageNet (IMAGENET) backgrounds.}} 
%     \label{fig:frac-results}
%     % \vskip -0.1in
% \end{figure*}
%
Figure \ref{fig:quantitative-results} shows explanation performance of individual sample-based importance maps produced by the selected XAI and baseline methods, across five models trained for each scenario-architecture parameterization, in terms of the $\mathrm{EMD}$ \red{and $\mathrm{IMA}$} metric\red{s}. 
Appendix \ref{app:quantitative-results} expands on the quantitative results of the main text, detailing results for all 16 methods studied and for our Precision metric.
In a few cases, performance tends to decrease as model complexity increases (from the simple LLR to the complex CNN architecture). 
One notable exception is for the RIGID scenario, where the CNN outperforms other models as expected. 
However, in this setting nearly all XAI methods are outperformed by a simple Laplace edge detection filter for correlated backgrounds results. 
\red{In this case, the discrepancy between the MLP and CNN performance is amplified for the $\mathrm{IMA}$ metric, with the CNN performing relatively better for a few XAI methods.}
The CNN also performs well in the case of the more-complicated IMAGENET backgrounds.

Within most scenario-architecture parameterizations, the performances of the studied XAI methods are relatively homogeneous, with a few exceptions. 
% Unlike the results shown by \citet{wilmingScrutinizingXAIUsing2022}, PatternNet does not outright outperform other methods for most non-linear parameterizations. 
% Most notably, PatternNet and PatternNet perform worse than GradSHAP, LRP, and LIME for the linear scenarios -- a direct contrast to the results of \citet{wilmingScrutinizingXAIUsing2022}. 
In most cases, correlated backgrounds (CORR) lead to worse explanation performance than their white noise (WHITE) counterparts, suggesting that suppressors in the smoothed background are difficult to distinguish from the class-dependent variables for most XAI methods.
\red{This effect can be most strongly observed when comparing RIGID WHITE to RIGID CORR for the $\mathrm{IMA}$ metric, suggesting that correlations in the background do indeed increase false positive attribution in model explanations.}

Baseline methods tend to perform similarly to one another. Interestingly, their performance is on par or even superior to various XAI methods in certain scenarios. 
Most notably, a simple Laplace edge detection filter outperforms nearly all other methods in the RIGID as well as the XOR scenarios, when used in combination with correlated backgrounds (CORR). 
\red{$\mathrm{IMA}$ results for baseline methods in the RIGID scenario show a lot less variance in the boxplots of Figure \ref{fig:frac-results} than for the $\mathrm{EMD}$ equivalents in Figure \ref{fig:emd-results}.}

\red{The results for the RIGID scenarios may be taken with a pinch of salt, as the high signal-to-noise ratios (SNRs) lead to highly salient tetrominoes in sample images. 
Notably, explanations produced for CNNs in this case tend to perform very well for both the $\mathrm{EMD}$ and $\mathrm{IMA}$ metrics compared to most results for any other model architecture and problem scenario.
While this problem itself (identifying a pattern with rotation and scaling invariance) is the most realistic of the four presented here, particularly when applied to CNNs, the high saliency of tetrominoes is perhaps not wholly akin to realistic problem settings, where the relative saliency of individual objects of interest is usually far lower.
The high saliency of the tetrominoes derives from our experimental choice to adjust SNRs to achieve a predefined minimal classification performance threshold, which required high SNR in this setting. An alternative approach could be to reverse this and fix the SNR for all scenarios and background types.}

\noindent \red{To revisit the stated questions from the start of Section \ref{sec:experiments}:}

\red{\textbf{1. Which XAI methods are best at identifying truly important features as defined by the sets $\mathcal{F}^+(\mathbf{x})$?}}

\red{The results show massive variability in performance for all methods across all problems and model architectures, so we cannot declare one specific ‘best’ method.
} 

\red{\textbf{2. Does explanation performance for each method remain consistent when moving from explaining a linear classification problem to problems with different degrees of non-linearity?}}

\red{Here we can see again that some methods vary in performance depending on the type of non-linearity (most perform better for MULT with the fixed position non-linearity than for RIGID), with a larger spread of $\mathrm{EMD}$ and $\mathrm{IMA}$ scores (seen in the size of boxes and whiskers of Figure \ref{fig:quantitative-results}) for non-linear scenarios than for LIN.} 

\red{The results for PatternNet and PatternAttribution \citep{kindermansLearningHowExplain2017} shown in the appendix (Figures \ref{fig:emd-full-results}, \ref{fig:frac-full-results} \ref{fig:precision-full-results}, \ref{fig:emd-full-results8x8}, and \ref{fig:precision-full-results8x8}) were proposed in part for solving the suppressor problem, and we can see how this is not necessarily always the case. 
These methods show strong performance for LIN as proposed, and as was seen in \citet{wilmingScrutinizingXAIUsing2022}, but do not look to generalize as well in most non-linear scenarios. 
Notably when the pattern signal is not in a fixed position (i.e., RIGID), these methods perform worse than when the signal is in a fixed position (i.e., MULT and XOR).
More specifically, they also look to learn the complete pattern signal (i.e., the tetromino shapes for both classes), so in the XOR case where both shapes are present and fixed in each sample, they do outright perform the best as one might expect.}

\red{\textbf{3. Does adding correlations to the background noise, through smoothing with the Gaussian convolution filter, negatively impact explanation performance?}}

\red{When looking at results from WHITE to CORR, we can spot a decrease in performance and increase in spread in most cases. 
This can be attributed to the fact that the imposed correlations (induced through Gaussian smoothing) between background pixels correlated with those overlapping with $\mathcal{F}^+$ cause background pixels to act as suppressor variables. 
One can control the strength of this effect by increasing/decreasing the strength of the Gaussian smoothing’s sigma parameter.}  

\red{\textbf{4. How does the choice of model architecture impact explanation performance?}}

\red{For LIN, explanation performance of all methods for all architectures is similar in most cases. 
When moving to non-linear scenarios, we can see little consistency in how architectures perform - the CNN can be seen to perform best in the RIGID case, but the MLP performs relatively better for the fixed tetromino position cases of MULT and XOR.
This can perhaps be explained by the CNN architecture tending itself well to rotation/translation invariance, whereas the properties of the MLP work better for a fixed-position ground-truth class-conditional distribution.}  

\red{We can also note that when multiple models present similar classification performance for a task, a user may assume or just not realize that explanation performance could be vastly different, as seen in the MLP vs CNN results of RIGID in Figure \ref{fig:quantitative-results}, and qualitatively in Figure \ref{fig:qualitative-results} across all architectures.}

\section{Discussion}\label{sec:discussion}
% Using carefully designed linear and non-linear benchmark data, we provide an objective quantitative evaluation of the accuracy of so-called XAI methods, which are methods attributing `importance' scores to input features of individual test samples. 
Experimental results confirm our main hypothesis that explanation performance is lower in cases where the class-specific signal is combined with a highly auto-correlated class-agnostic background (CORR) compared to a white noise background (WHITE). 
% This was tested for additive and multiplicative combinations of signal and background. 
The difficulty of XAI methods to correctly highlight the truly important features in this setting can be attributed to the emergence of suppressor variables. 
% Our results, therefore, go beyond the results for the linear case shown by \citet{wilmingScrutinizingXAIUsing2022}. 
Importantly, the misleading attribution of importance by an XAI method can lead to misinterpretations regarding the functioning of the predictive model, which could have severe consequences in practice. 
Such consequences could be unjustified mistrust in the model's decisions, unjustified conclusions regarding the features related to a certain outcome (e.g., in the context of medical diagnosis), and a reinforcement of such false beliefs in human-computer interaction loops.

We have also seen that when multiple ML architectures can be used interchangeably to appropriately solve a classification problem -- here with classification accuracy required to be above 80\% -- they may still produce disparate explanations. 
% This discrepancy is especially apparent for CNNs, which failed to produce meaningful explanations on numerous occasions. 
Architectures not only differed with respect to the selection of pixels within the correct set of important features, but also showed different patterns of false positive attributions of importance to unimportant background features. 
If one cannot produce consistent and sensical results for multiple seemingly appropriate ML architectures, the risk of model mistrust may be especially pronounced.
% \subsection{Limitations}

% It has been observed by \citeauthor{arrasCLEVRXAIBenchmark2022} that there may be a link between classification performance and explanation performance. 
% Whilst we have tried to mitigate this in our experimental setup, models trained over the CNN architecture tend to perform worse than LLR and MLP equivalents for a given $\alpha$. One could also consider fixing test accuracy instead, however the reliance on testing over data generated with different values of $\alpha$ has its own limitations.

A recent survey showed that one in three XAI papers evaluate methods exclusively with anecdotal evidence, and one in five with user studies \citep{nauta2023anecdotalevidence}.
Other work in the field tends to focus on secondary criteria (such as stability and robustness \citep{rosenfeldRisksInvariantRisk2021, hedstroemQuantusExplainableAI2022}) or subjective or potentially circular criteria (such as fidelity and faithfulness \citep{gevaertEvaluatingFeatureAttribution2022, nauta2023anecdotalevidence}). 
\red{It was recently shown in \citet{wilmingTheoreticalBehavior2023} that faithfulness as a concept can be treated as an XAI method in itself, and when done so is also prone to the attribution of arbitrarily high importance to suppressor variables.}
We \red{therefore} doubt that such \red{secondary} validation approaches can fully replace metrics assessing objective notions of `correctness' of explanations, considering that XAI methods are widely intended to be used as means of quality assurance for machine learning systems in critical applications. 
Thus, the development of specific formal problems to be addressed by XAI methods, and the theoretical and empirical validation of respective methods to address specific problems, is necessary.
In practice, a stakeholder may often (explicitly or implicitly) expect that a given XAI method identifies features that are truly related to the prediction target. 
In contrast to other notions of faithfulness, this is an objectively quantifiable property of an XAI method, and we here propose various non-linear types of ground-truth data along with appropriate metrics to directly measure explanation performance according to this definition. 
While our work is not the first to provide quantitative XAI benchmarks \citep[see,][]{tjoaQuantifyingExplainabilitySaliency2020, li2021experimental, zhouFeatureAttributionMethods2022,arrasCLEVRXAIBenchmark2022,gevaertEvaluatingFeatureAttribution2022,agarwalOpenXAITowardsTransparentEvaluation2022}, our work differs from most published papers in that it allows users to quantitatively assess potential misinterpretations caused by the presence of suppressor variables in data.

% Li et. al (2020) provide an “exhaustive experimental study” based on the “status quo” of XAI performance metrics, and while they mention a precision-based metric known as IoSR as the only objective performance metric, they dedicate a significant portion of the paper

% \subsection{Limitations}\label{sec:limitations}
One potential limitation of \red{the $\mathrm{EMD}$ metric} is the strictness of limiting the ground truth feature set $\mathcal{F}^+$ to the specific pixels of tetrominoes $\red{\boldsymbol{a}}^{\text{T/L}}$ compared to, say, the set of features outlining $\red{\boldsymbol{a}}^{\text{T/L}}$. 
Alternative definitions of $\mathcal{F}^+$ could be conceived to more flexibly adapt to different potential `explanation strategies'. 
\red{Figure \ref{fig:emd-intuition} in the appendices outlines four `explanation strategies' and how the $\mathrm{EMD}$ metric varies with each. 
Notably, an `outline' explanation performs worse than an explanation highlighting a subset of $\mathcal{F}^+$. 
This highlights two interesting features of our novel metric. 
Firstly, a strongly performing `subset' explanation shows that $\mathrm{EMD}$ does not penalize false negatives (not attributing high importance to some truly important features) as harshly as Precision and other `top-k' metrics do. 
Secondly, the `outline' explanation functions in a presumably similar way to some model-ignorant edge detection methods, and performs the worst of any explanation strategy shown in Figure \ref{fig:emd-intuition}. 
Yet, we have shown such edge detection methods to be capable of outperforming many XAI methods in some problem scenarios.
Our $\mathrm{IMA}$ metric also complements this potential limitation of $\mathrm{EMD}$, where it does not matter if the attribution of importance to features of $\mathcal{F}^+$ is spread across all features, or just more intensely attributed to a subset. 
This metric directly measures false positive attribution of importance to features outside of $\mathcal{F}^+$, and assists the user in understanding the role that suppressors play in model explanations.}
%A future modification to the $\mathrm{EMD}$ ground metric to set the cost of optimal transport between features within $\mathcal{F}^+$ could provide the best of both worlds between both metrics. 
%This would not penalize `subset'-style explanations for needing to evenly distribute importance across all features of $\mathcal{F}^+$, whilst also not penalizing explanations highlighting features in close proximity to the ground truth (i.e., `outline'-style explanations) as is done for $\mathrm{IMA}$ and Precision.}

% While our $\mathrm{EMD}$ metric would favor an explanation pertaining to the latter definition over a nonsensical explanation, an alternative metric with a specific proximity penalty could be incorporated into the benchmark at a later stage.
While we compare a total of 16 XAI methods, the space of possible neural network architectures is too vast to be represented; therefore we only compared one MLP and one CNN architecture here. However, our experiments hopefully serve as a showcase for our benchmarking framework, which can be easily extended to other architectures. 
Finally, our framework serves much needed validation purposes for methods that are conceived to themselves play a role in the quality assurance of AI. 
As such, we expect that the benefits of our work far outweigh potential negative implications on society, if any. 
A possible risk, even if far-fetched, would be that one may reject a fit-for-purpose XAI method based on empirical benchmarks such as ours, which do not necessarily reflect the real-world setting and may hence be too strict.  

\section{Conclusion}
We have used a data-driven generative definition of feature importance to create synthetic data with well-defined ground truth explanations, and have used these to
provide an objective assessment of XAI methods when applied to various classification problems. 
Furthermore, we have defined new quantitative metrics of explanation performance and demonstrated that many popular XAI methods do not behave in an ideal way when moving from linear to non-linear scenarios. 
Our results show that XAI methods can even be outperformed by simple model-ignorant edge detection filters in the RIGID use case, in which the object of interest is not located in a static position. 
Finally, we show that XAI methods may provide inconsistent explanations when using different model architectures under equivalent conditions.  
% Future work will be to use our benchmarks to develop improved XAI methods as well as further develop performance benchmarks and metrics to assess their quality.
Future work will be to develop dedicated performance benchmarks in more complex and application-specific problem settings such as medical imaging. 

\section{Declarations}
\textbf{Funding} - This result is part of a project that has received funding from the European Research Council (ERC) under the European Union’s Horizon 2020 research and innovation programme (Grant agreement No. 758985), the German Federal Ministry for Economic Affairs and Climate Action (BMWK) within the ``Metrology for Artificial Intelligence in Medicine (M4AIM)'' program in the frame of the ``QI-Digital'' initiative, and the Heidenhain Foundation in the frame of the Junior Research Group ``Machine Learning and Uncertainty''.

\noindent\textbf{Conflicts of interest/Competing interests} - The authors declare no conflicts of interest/competing interests.

\noindent\textbf{Availability of data and material} - All data used here can be generated using the provided code.

\noindent\textbf{Code availability} - \url{https://github.com/braindatalab/xai-tris}

\noindent\textbf{Ethics approval} - Not applicable.

\noindent\textbf{Consent to participate} - Not applicable.

\noindent\textbf{Consent for publication} - Not applicable.

\noindent\textbf{Author contributions} - All authors contributed to the study conception and design. Material preparation, data collection and analysis were performed by Benedict Clark. The first draft of the manuscript was written primarily by Benedict Clark, and all authors commented on and edited all previous versions of the manuscript. All authors read, edited, and approved the final manuscript.

% \section*{References}

% References follow the acknowledgments. Use unnumbered first-level heading for
% the references. Any choice of citation style is acceptable as long as you are
% consistent. It is permissible to reduce the font size to \verb+small+ (9 point)
% when listing the references.
% Note that the Reference section does not count towards the page limit.
% \medskip

% {
% \small

% [1] Alexander, J.A.\ \& Mozer, M.C.\ (1995) Template-based algorithms for
% connectionist rule extraction. In G.\ Tesauro, D.S.\ Touretzky and T.K.\ Leen
% (eds.), {\it Advances in Neural Information Processing Systems 7},
% pp.\ 609--616. Cambridge, MA: MIT Press.

% [2] Bower, J.M.\ \& Beeman, D.\ (1995) {\it The Book of GENESIS: Exploring
%   Realistic Neural Models with the GEneral NEural SImulation System.}  New York:
% TELOS/Springer--Verlag.

% [3] Hasselmo, M.E., Schnell, E.\ \& Barkai, E.\ (1995) Dynamics of learning and
% recall at excitatory recurrent synapses and cholinergic modulation in rat
% hippocampal region CA3. {\it Journal of Neuroscience} {\bf 15}(7):5249-5262.
% }

\bibliography{xai_better}
% \bibliographystyle{icml2023}

%%%%%%%%%%%%%%%%%%%%%%%%%%%%%%%%%%%%%%%%%%%%%%%%%%%%%%%%%%%%
% \newpage
% ~\newpage
% ~\newpage

% \afterpage{
% \pagestyle{empty}
% \newpage~\newpage~\newpage~
% %\setcounter{page}{2}
% }

\newpage
\begin{appendices}
    \section{MLJ Contribution Information Sheet}

\paragraph{What is the main claim of the paper? Why is this an important contribution to the machine learning literature?}

We claim that many post-hoc explanation methods consistently and reproducibly highlight certain input features that have no statistical dependency to the target variable predicted by the model. 
The existence of such so-called suppressor variables, and the false positive attribution of such variables as important, can lead to severe misinterpretations, which raises concerns regarding the correctness and utility of ‘explanations’ provided by explanation methods.

We create benchmark image datasets for one linear and three non-linear classification scenarios, in which the important class-conditional features are known by design. These scenarios are based on different types and combinations of tetrominoes \citep{golombPolyominoesPuzzles1996}, overlaid on one of three types of noisy backgrounds. One of these background types, white noise smoothed by a Gaussian filter, induces the presence of suppressor variables through the correlation of background pixels overlapping the tetromino with those just of the noisy background. In all cases, ground truth explanations are explicitly known through the location of the tetrominoes in the sample.  

We develop novel performance metrics, one based on the Earth mover's distance of transforming the `energy' of a given explanation into the ground truth explanation, and use this to show that in many cases, the presence of induced suppressor variables hinders explanation performance for many popular XAI methods. Another metric directly measures the false positive attribution of model explanations through the proportion of importance attributed to ground truth features over the total attribution of the explanation. These two metrics complement each other well.

Through our experimental results we draw other conclusions, including that explanations produced for different equally performing ML architectures can be very inconsistent. We show that popular explanation methods are sometimes unable to outperform random performance baselines and edge detection methods. We highlight that secondary metrics such as faithfulness are currently not sufficient to assess ML explanation quality compared to objective metrics focused on the `correctness' of explanations, such as those presented here.

The importance of these claims is that machine learning model explanations are prone to misinterpretation under such inconsistencies. For example, one may assume that equally performing models would produce equally performing explanations, however this is not always true. One may have chosen a particular architecture based on other properties of it, and end up with misleading or nonsensical explanations.
We necessitate that for XAI to be deployed in high-stakes fields, such risks should be mitigated. Our approach is a rigorous and objective evaluation of the performance of current explanation methods, which can lead to the development of stronger and more reliable methods in the future.

\paragraph{What is the evidence you provide to support your claim? Be precise.}

We conduct an extensive set of empirical experiments across 4 image classification problem scenarios, 3 background types, 3 model architectures, 16 explanation methods, 4 performance baselines, and 3 metrics. 
We carefully construct the important class-conditional features in each problem, which can serve as ground truth explanations.
We assess many popular post-hoc XAI methods and quantify their `explanation performance' using metrics from signal detection theory such as Earth mover's distance, $\mathrm{IMA}$, and precision, and show that such methods attribute importance to suppressor variables and can lead to misleading interpretations.

Through our experimental results we observe behavior including that explanations produced for different equally performing ML architectures can be very inconsistent. We show that popular explanation methods are sometimes unable to outperform random performance baselines and edge detection methods for our developed performance metrics. We discuss, using related literature, that secondary metrics such as faithfulness are currently not sufficient to assess ML explanation quality compared to objective metrics focused on the `correctness' of explanations, such as those presented here.

\paragraph{What papers by other authors make the most closely related contributions, and how is your paper related to them?}

Several works in the XAI field have moved towards quantitative evaluation of XAI methods using ground truth data \citep{tjoaQuantifyingExplainabilitySaliency2020, li2021experimental, zhouFeatureAttributionMethods2022, arrasCLEVRXAIBenchmark2022, gevaertEvaluatingFeatureAttribution2022, agarwalOpenXAITowardsTransparentEvaluation2022}. 
However, these studies are limited in the extent to which they perform quantitative assessment, and many such studies do not construct their benchmark datasets in a way that realistic correlations between class-dependent and class-agnostic features (i.e., the foreground/object in an image vs. the background) are included. 
In practice, these correlations can give rise to features acting as suppressor variables. These works do not focus on such variables and our previous work is the only such work to do so. 

\citet{wilmingScrutinizingXAIUsing2022}, published in ECML 2022, took a similar approach to that shown here, yet focused on a linear problem for one model architecture, and did not make use of random performance baselines to compare XAI methods to.
\citet{wilmingTheoreticalBehavior2023} also looked into quantifying explanation performance in the presence of suppressors using a two-dimensional linear example, however the focus there was on analytically deriving the exact influence of suppressors on produced explanations.

\paragraph{Have you published parts of your paper before, for instance in a conference? If so, give details of your previous paper(s) and a precise statement detailing how your paper provides a significant contribution beyond the previous paper(s).}

The content of this paper is entirely original. Some ideas discussed in this paper have already been voiced in our prior work \citep{haufeInterpretationWeightVectors2014, wilmingScrutinizingXAIUsing2022, wilmingTheoreticalBehavior2023}.
However, our current paper goes beyond these through focusing on an extensive set of empirical experiments across 4 image classification problem scenarios, 3 background types, 3 model architectures, 16 explanation methods, 4 performance baselines, and 3 metrics.

\paragraph{Suggested Reviewers}
Pieter-Jan Kindermans (pikinder@google.com): Author of PatternNet and PatternAttribution.

\noindent Moritz Grosse-Wentrup (moritz.grosse-wentrup@univie.ac.at): Expert in XAI and causality.

\noindent Max Welling (M.Welling@uva.nl): Esteemed machine learning expert with interest in XAI.

\noindent Robert Jenssen (robert.jenssen@uit.no): Professor of machine learning with track record in XAI.

\clearpage
\newpage

\section{Appendix}

The authors confirm that we bear all responsibility in case of violation of rights of any kind in the data and results shown in this work.

\subsection{ImageNet}\label{sup:imagenet}
We sample data from the ImageNet-1k subset \citep{deng2009imagenet}, following the license specified here \url{https://image-net.org/download.php}.

In the ImageNet-1k subset, there are only three people categories (scuba diver, bridegroom, and baseball player) included in the 1,000 classes, versus 2,832 people categories in the full set. 
There is also the possibility of people-related images co-existing in images of other classes, which has been noted \citep{prabhu2020large}.
Data from these classes can be discarded if necessary.

Alternatives can be used directly as a background type here to replace ImageNet, for example PASS \citep{asano2021pass}, published in the NeurIPS Datasets and Benchmarks track in 2021. 
This ImageNet replacement dataset only contains images with a CC-BY license, as well as containing no images of humans. 
Replacement of ImageNet images in our work is as simple as placing images in the respective folder for the data generation step to handle, following the instructions outlined in the next sub-section and the corresponding GitHub repository. 

\subsection{Code and Data}\label{sup:code-data}
All code for generating data and performing model training and XAI analysis is available on GitHub: \url{https://github.com/braindatalab/xai-tris}. 
There, we provide instructions on how to run each step of the analysis pipeline as well as detailing corresponding configuration fields.

To download the ImageNet data, we made an account and agreed the license terms on \url{https://huggingface.co/datasets/imagenet-1k} and subsequently downloaded the data. 
Here, we used the validation set as the $N=50,000$ set suited the volume requirement for our analysis.
We of course advise anyone planning to do similar analysis on a model pre-trained with ImageNet data to use the $N=100,000$ test set instead.

Each $N=40,000$ dataset generated for a given classification scenario and background type pair is $1.52$ GB in size. 
For the lower-dimensional $8 \times 8$-px data and experiments shown in supplementary materials Section \ref{sup:8by8}, generating $N=10,000$ datasets for all eight scenario and background type pairs is around $62$ MB in total size, and was combined in one file due to this much lower volume requirement.
Each scenario's dataset is saved as a file \lstinline|SCENARIO_JdKp_$\alpha$_BACKGROUND.pkl| containing a python dictionary 
\begin{lstlisting}
    {SCENARIO_JdKp_$\alpha$_BACKGROUND : DataRecord(...)},
\end{lstlisting}
where \lstinline|SCENARIO={linear, multiplicative, translations_rotations, xor}| and \lstinline|BACKGROUND={white, correlated, imagenet}|. Image scale \lstinline|Jd={1,8}d| is the scaling of the image dimensionality $d$ from the original $8 \times 8$-px images to the $64 \times 64$-px images shown in the main text, pattern scale \lstinline|Kp={1,4,8}p| is the scaling of the tetromino pattern (width in pixels), and $0.0 \leq \alpha \leq 1.0$ parameterizes the signal-to-noise ratio.

\lstinline|DataRecord| is a Python \lstinline|namedtuple()| collection specified as
\begin{lstlisting}
    DataRecord = namedtuple(`DataRecord', `x_train y_train x_val y_val x_test y_test masks_train masks_val masks_test').
\end{lstlisting}
Each field can be accessed programmatically via the name, for example \lstinline|DataRecord.x_test| returns the test data $\mathbf{x}_{\text{test}}$ of the dataset. 
The \lstinline|masks| fields are the tetromino pattern masks which form the ground truth for explanations.

\subsection{Compute} \label{sup:compute}
Experiments were run on a cluster consisting of four Nvidia A40 GPUs, where each model training took roughly between three and twenty minutes to complete, depending on architecture. 
Time estimation for running XAI methods is more rough to calculate and depends on each method, but in total for all models and methods for a given scenario's $N=2,000$ test set, this took between 24 and 48 hours of compute time per GPU on the cluster.
Quantitative analysis took roughly a further 24 hours of compute per scenario on a cluster of AMD EPYC 7702 CPUs, with six threads used for each of the 12 scenarios.

Due to smaller compute requirements, we can also recommend that if one wants to explore the code and data with smaller compute requirements, the $8 \times 8$-px data shown in supplementary materials Section \ref{sup:8by8} is also representative of a strong benchmark for XAI methods. 
Code and instructions to run it have also been provided in the GitHub repository linked in the above supplementary materials Section \ref{sup:code-data}.

\subsection{Data}
\begin{figure}[!ht]
  \centering
    \includegraphics[width=\textwidth]{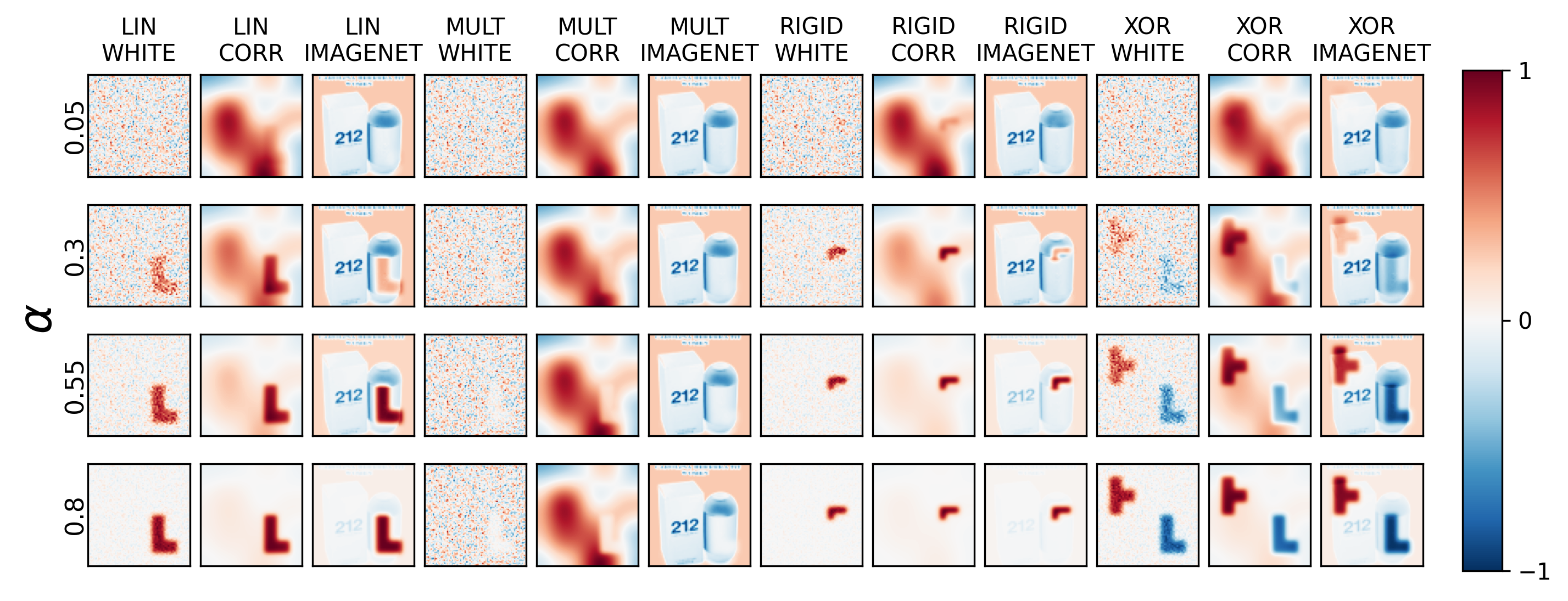}
    \caption{Examples of generated data samples for each scenario, showing how a generated sample of Class \#0 (where y=0)  for each scenario varies across four signal-to-noise ratios (SNRs) $\alpha$.}
    \label{fig:snr_plot}
  % \vskip -0.1in
\end{figure}

Here, we expand on Figure \ref{fig:data_plot} with Figure \ref{fig:snr_plot}, which shows an example of each scenario across four choices of signal-to-noise ratio (SNR), parameterized by $\alpha$.
% \begin{figure}[ht]
%     \centering
%     \includegraphics[width=\textwidth]{figures/snr_plot.png}
%     \caption{Example of how one generated sample for each scenario varies across four different SNRs $\alpha$. As $\alpha$ increases, tetromino patterns for the additive scenarios LIN, RIGID, and XOR show stronger positive and negative values. For the multiplicative scenario MULT, the tetromino pattern is pulled towards zero as $\alpha$ increases.}
%     \label{fig:snr_plot}
% \end{figure} 

% \begin{figure}[!ht]
%     \centering
%     \begin{subfigure}{\textwidth}
%         \centering
%         \includegraphics[width=\linewidth]{figures/snr_plot.png}
%         \caption{One generated sample of Class \#0 (where y=0) for four different SNRs $\alpha$.}
%     \label{fig:snr_plot_subfig}
%     \end{subfigure} 
%     %
%     \begin{subfigure}{\textwidth}
%         \raggedleft
%         \includegraphics[width=0.98\linewidth]{figures/data_plot_corrected.png}
%         \caption{Two generated samples of each class per scenario, with 4px-thick tetrominoes in the RIGID case.}
%         \label{fig:data_plot_corrected}
%     \end{subfigure}
%     \caption{Examples of generated data samples for each scenario, showing how an example for each scenario varies across four signal-to-noise ratios (SNRs) $\alpha$ (top), and an updated version of Figure \ref{fig:data_plot} with 4px-thick tetrominoes in the RIGID case as was used in subsequent experiments.}
%     \label{fig:snr_plot}
%     \vskip -0.1in
% \end{figure}

  % \caption{Examples of generated data samples for each scenario, showing differences between samples of the two classes (top) and how an example for each scenario varies across four signal-to-noise ratios (SNRs) $\alpha$ (bottom).}

\subsection{Explanation Methods and Model Training}\label{app:methods-training}
% \subsubsection{Explanation Methods}\label{app:methods}
Here, we detail the full suite of 16 XAI methods used in our analysis, with a brief description along with the reference and any parameterization details.
In the main text, we focus on XAI methods available with the Captum \citep{kokhlikyanCaptumUnifiedGeneric2020} framework for explaining PyTorch models. 
We also make use of methods available in the iNNvestigate \citep{alberINNvestigateNeuralNetworks2018} library, through training equivalent models for the Keras framework. 

\begin{longtable}{  p{0.2\textwidth}  p{0.41\textwidth}  p{0.18\textwidth} p{0.11\textwidth}  }
\caption{XAI Methods used with a brief description of each method and the implementation details, including the software framework used and any specific parameterization including the baseline input used, if applicable.}\label{tab:xai_methods_table}\\
\toprule
\vspace{5pt}
\textbf{XAI Method} & \textbf{Description} & \textbf{Implementation Framework, Parameterization} & \textbf{Reference}\\ \midrule
Permutation Feature Importance (PFI)
& Measures the change in prediction error of the model after permuting each feature's value
& Captum, Default
& \cite{fisherAllModelsAre2019} 
\\
Integrated Gradients
& Computes gradients along the path from a baseline input to the input sample, and cumulates these through integration to form an explanation
& Captum, Default, Zero input baseline
& \cite{sundararajanAxiomaticAttributionDeep2017} 
\\
Saliency
& Computes the gradients with respect to each input feature
& Captum, Default
& \cite{Simonyan14a} 
\\
Guided Backpropagation
& Computes the gradient of the output with respect to the input, but ensures only non-negative gradients of ReLU functions are backpropagated
& Captum, Default
& \cite{springenbergStrivingSimplicityAll2015} 
\\
Guided GradCAM
& Computes the element-wise product of guided backpropagation attributions with respect to a class-discriminative localization map in the final convolution layer of a CNN. This produces a coarse importance map for the target class as an explanation, the same size as the convolutional feature map, rather than pixel-wise over the whole image
& Captum, Default
& \cite{selvarajuGradCAMVisualExplanations2017} 
\\
Deconvolution
& Uses a Deconvolutional network to map features to pixels. An explanation is produced by computing the gradient of the target output, only backpropagating non-negative gradients of ReLU functions
& Captum, Default
& \cite{zeilerVisualizingUnderstandingConvolutional2014} 
\\
DeepLift
& Compares the difference between the activation of each neuron and its `reference activation', and produces an explanation based on this difference
& Captum, Default, Zero input baseline
& \cite{shrikumarLearningImportantFeatures2017} 
\\
Shapley Value Sampling
& Approximates Shapley values by repeatedly sampling random permutations of input features and calculating the contribution of each feature to the prediction. An explanation is produced across an average of many samplings 
& Captum, Default, Zero input baseline
& \cite{castroPolynomialShapleySampling2009} 
\\
Gradient SHAP
& Approximates Shapley values by computing the expected values of gradients when randomly sampled from the distribution of baseline samples
& Captum, Default, Zero input baseline
& \cite{lundbergUnifiedApproachInterpreting2017} 
\\
Kernel SHAP
& Approximates Shapley values through the use of LIME, setting the loss function, weighting kernel, and regularization term in accordance with the SHAP framework 
& Captum, Default, Zero input baseline
& \cite{lundbergUnifiedApproachInterpreting2017} 
\\
Deep SHAP
& Approximates Shapley values through the use of DeepLift. Computes the DeepLift attribution for each input sample with respect to each baseline sample, in accordance with the SHAP framework
& Captum, Default, Zero input baseline
& \cite{lundbergUnifiedApproachInterpreting2017}  
\\
Locally-interpretable Model Agnostic Explanations (LIME)
& Learns a linear surrogate model locally to an individual prediction, perturbing and weighting the dataset in the process, and then builds an explanation by interpreting this local model  
& Captum, Default
& \cite{ribeiroWhyShouldTrust2016} 
\\
Layer-wise Relevance Propagation (LRP)
& Propagates the model output back through the network as a measure of relevance, decomposing this score for each model in each layer based on their trained weight and activation
& Captum, Default
& \cite{bachPixelWiseExplanationsNonLinear2015} 
\\
Deep Taylor Decomposition (DTD)
& Applies a Taylor decomposition from a specified root point to approximate the sub-functions of a network, building explanations by applying this backward from the network output to input variables   
& iNNvestigate, Default
& \cite{montavonExplainingNonLinearClassification2017} 
\\
PatternNet
& Estimates activation patterns per neuron through signal estimator $S_{\mathbf{a}+-}$ and back-propagates this through the network. The explanation is given as a projection of the signal in input space
& iNNvestigate, Default
& \cite{kindermansLearningHowExplain2017} 
\\
PatternAttribution
& Utilises the theory of PatternNet to estimate the root point of the data for DTD, and yields the attribution $\mathbf{w} \odot \mathbf{a}_+$ for weight vector $\mathbf{w}$ and positive activation patterns $\mathbf{a}_+$. The explanation is given as the neuron-wise contribution of the signal to the classification score
& iNNvestigate, Default
& \cite{kindermansLearningHowExplain2017} 
\\
\bottomrule
\end{longtable}

\begin{figure}[!ht]
    \centering
    \includegraphics[width=\textwidth]{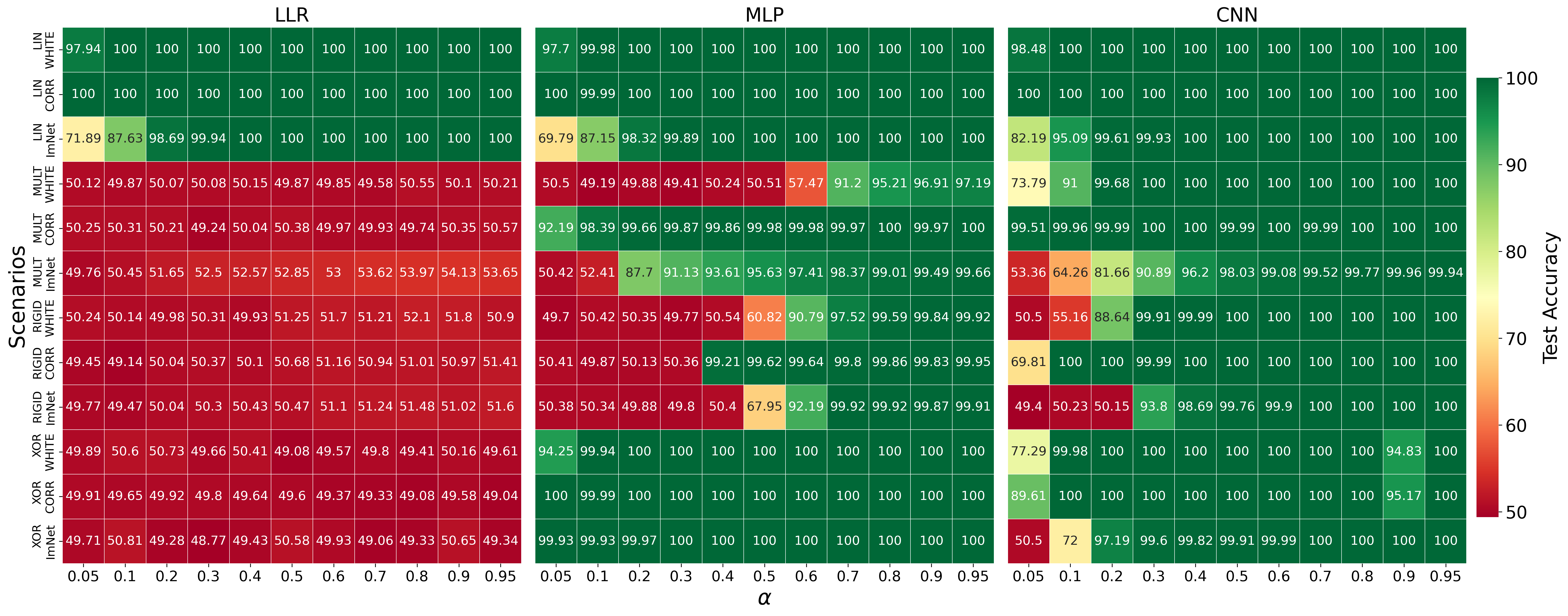}
    \caption{Average test accuracy over 10 model trainings for each problem scenario and model architecture, for a fixed range of signal-to-noise ratios (SNRs). As expected, the Linear Logistic Regression (LLR) model cannot perform above chance level for non-linear scenarios. The Convolutional Neural Network (CNN) outperforms the Multi-Layer Perceptron (MLP) for the RIGID (translations and rotations of tetrominoes) scenarios as expected, perhaps due to the invariance under these properties for this architecture. }
    \label{fig:training-results}
\end{figure}

\subsection{Earth Mover's Distance}

\begin{figure}[!ht]
    \centering
    \includegraphics[width=\textwidth]{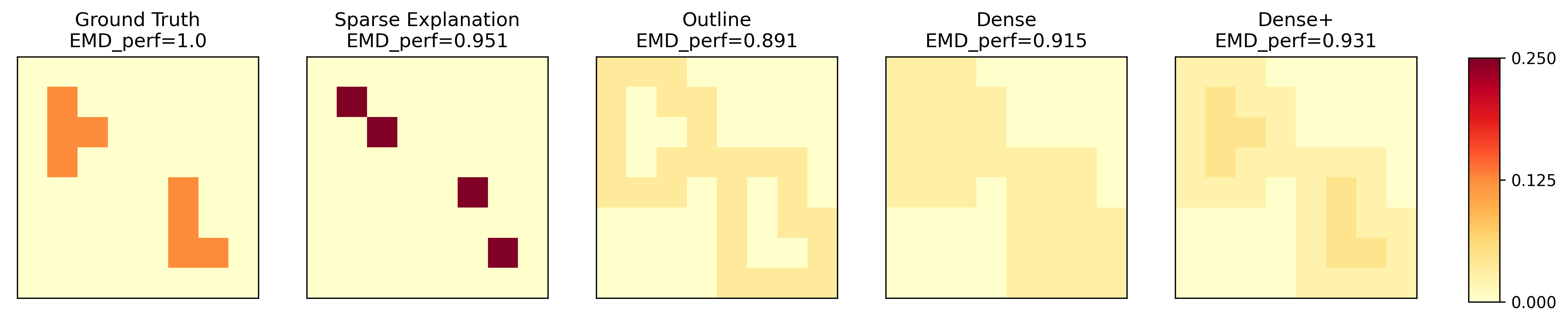}
    \caption{\red{$\mathrm{EMD}$ scores for the $8 \times 8$ ground truth as well as four `explanation strategies'. Here, we can see that the $\mathrm{EMD}$ metric does not penalize an explanation highlighting a subset of truly important features compared to an explanation highlighting the outline of the ground truth. This shows that the $\mathrm{EMD}$ penalizes false negatives (not attributing high importance to truly important features) less than a `top-k' metric like Precision would. The `outline' strategy in the third column produces an explanation presumably similar to a model-ignorant edge detector, which has the lowest $\mathrm{EMD}$ score of the strategies shown, yet we have shown such edge detectors can outperform many XAI methods in some problem scenarios.}}
    \label{fig:emd-intuition}
\end{figure}

% \begin{figure}[!ht]
%     \centering
%     \includegraphics[width=\textwidth]{figures/linesearch_plot.png}
%     \caption{Average test accuracy over 10 model trainings for each classification problem and background type across the three model architectures. Here, a more fine-grained range of 10 signal-to-noise ratios (SNRs) was chosen for each problem-background pair so to find a `sweet spot' of model performance above 80\% test accuracy, compared with the large gaps of performance between SNRs seen in Fig.~\ref{fig:training-results}. The result of this was used to guide the choice of SNRs shown in Table \ref{tab:chosen-snrs}.}
%     \label{fig:line-search}
% \end{figure}

\subsection{Explanation Performance}\label{app:xai-results}
This section further elaborates results of our experiments on validating the performance of XAI methods.
In Figures \ref{fig:emd-full-results} and \ref{fig:precision-full-results} we also show methods available in the iNNvestigate \citep{alberINNvestigateNeuralNetworks2018} library, through training equivalent models for the Keras framework.
We note that there were some issues in convergence for CNN models for the XOR scenarios with the required Keras framework, even under seemingly equivalent conditions such as fixed random seeds and He-normal weight initialization. 
Our model architectures have been chosen as a showcase of the datasets and benchmarks of this work, and other architectures may have better or worse performance on the same XAI methods, but this was not a focus of this work.
As such, we do not show the corresponding results for these methods (PatternNet, PatternAttribution, Deep Taylor Decomposition) in the XOR-CNN problem setting, so to promote a fair comparison of methods.

\subsubsection{Qualitative Results}\label{app:qualitative-results}
\begin{figure}[!ht]
    \centering
    \includegraphics[width=0.9\textwidth]{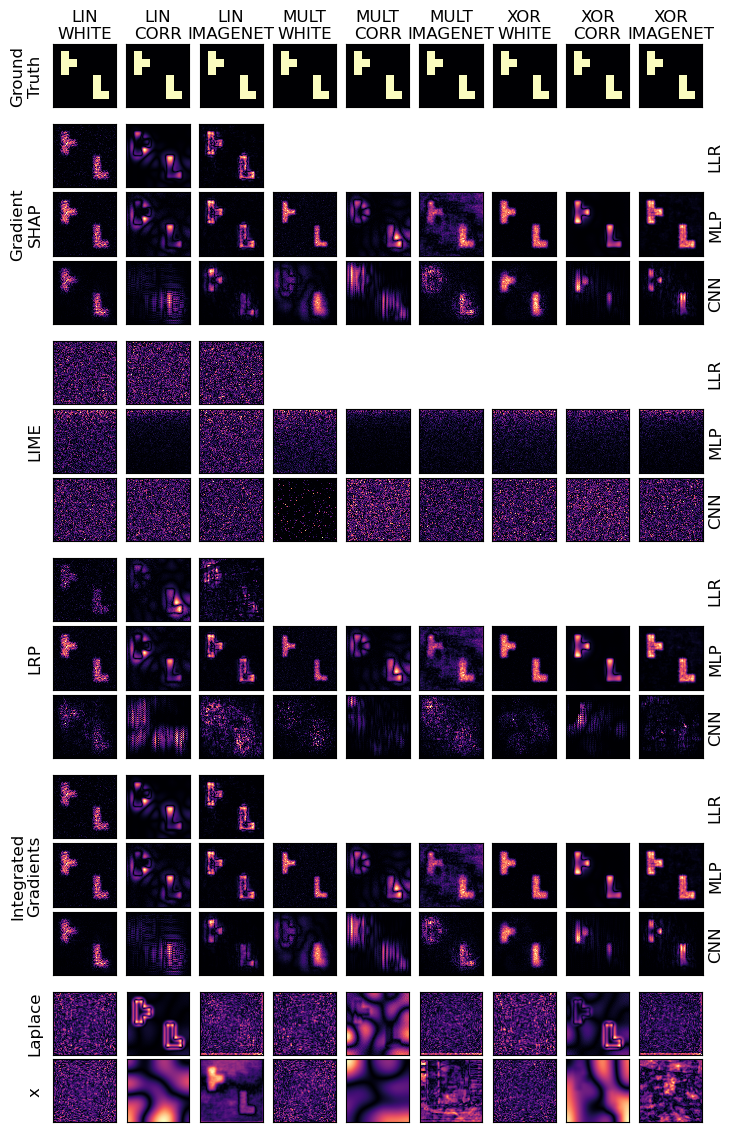}
    \caption{Absolute-valued global importance maps calculated as the mean importance value over all correctly predicted samples, for selected XAI methods and baselines. RIGID scenarios involving translations and rotations of the tetromino signal pattern are not included as they have no fixed ground truth position. CORR scenarios with correlated background can be seen to produce noisier global importance maps, suggesting that this setting induces suppressor variables in the background, which are difficult for XAI methods to distinguish from the true signal pattern. Results for the ImageNet background also tend to show noisier global explanations, suggesting that the complicated and variable features of this background type present a challenge to the models and corresponding XAI methods. LIME fails to produce any meaningful explanations yet again, suggesting an issue with this scale of image. The results of supplementary materials Section \ref{sup:8by8} show better performance for LIME with the smaller $8 \times 8$-px image benchmark.}
    \label{fig:qualitative-results-global}
\end{figure}

In Figure \ref{fig:qualitative-results-global}, we can see absolute-valued global importance maps for selected XAI methods and baselines, calculated as the mean importance value over all correctly predicted samples. RIGID scenarios involving translations and rotations of the tetromino signal pattern are not included as they have no fixed ground truth position.
% \todo{NOTE: it would be hard to produce a plot of this format with all methods included, but I can produce a plot of global saliency maps for each model individually, so to have 3 plots including all methods}x

\subsubsection{Quantitative Results}\label{app:quantitative-results}
\begin{figure}[!ht]
    \centering
    \includegraphics[width=\textwidth]{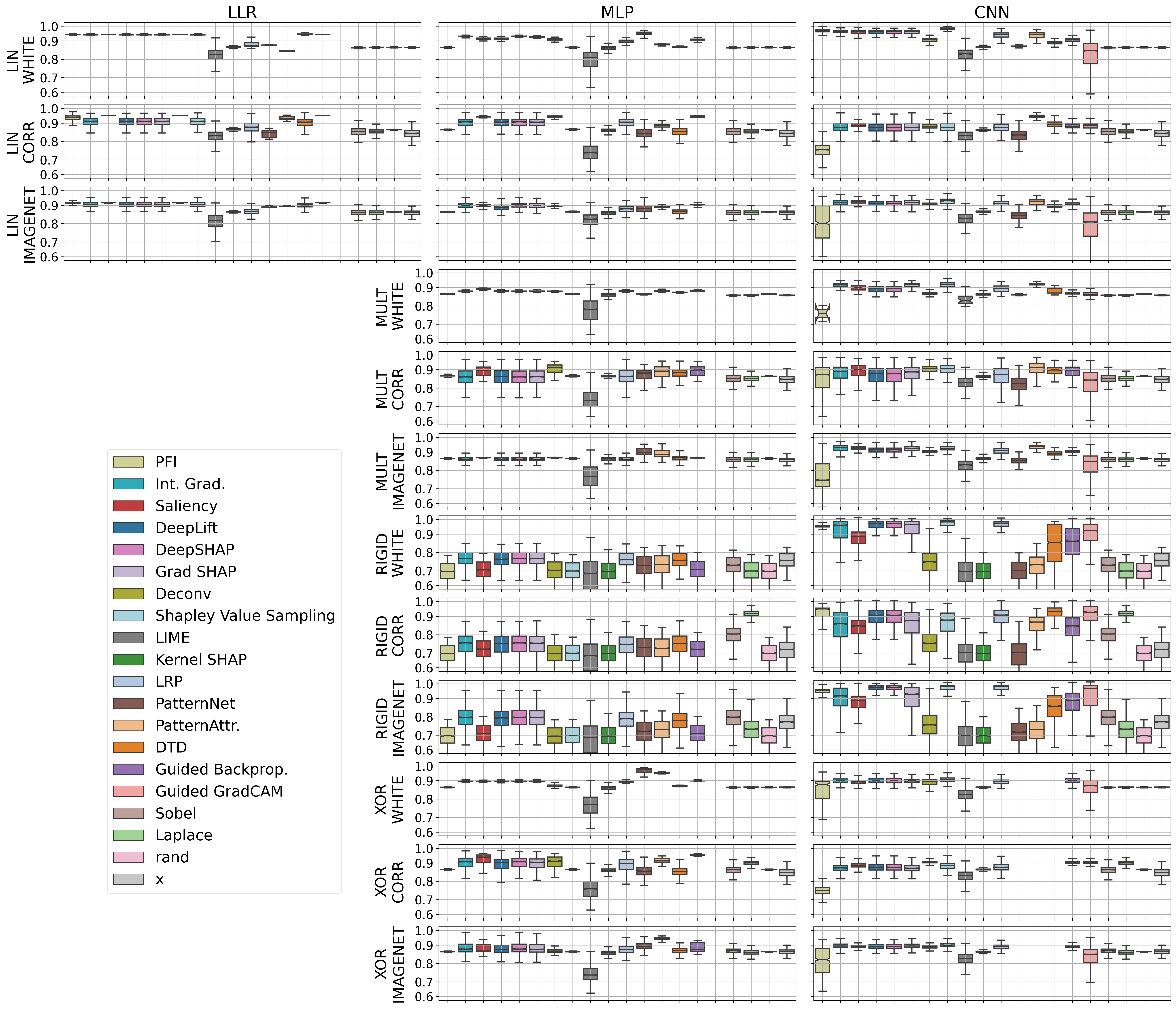}
    \caption{$\mathrm{EMD}$ metric based on the Earth Mover's Distance (EMD) for every XAI method tested, separated by model architecture and depicted as boxplots of median and quartile performance scores. Guided GradCAM is only implemented for CNN architectures, and Keras models required for PatternNet, PatternAttribution, and Deep Taylor Decompostion (DTD) struggled to converge for the XOR scenarios as stated above, so these are excluded from the corresponding sub-plots. Some methods see a drop in explanation performance as model complexity increases, from the Linear Logistic Regression (LLR) model to a Convolutional Neural Network (CNN). In the RIGID CORR case,  the model-ignorant Laplace filter outright performs the best for explanations of MLP decisions and nearly so for the CNN. SHAP variants DeepSHAP, GradSHAP, and Shapley Value Sampling perform very similarly to one another in most cases across all model types, despite being formulated to target particular problems. No XAI method performs outright the best across all scenarios.}
    \label{fig:emd-full-results}
\end{figure}

\begin{figure}[!ht]
    \centering
    \includegraphics[width=\textwidth]{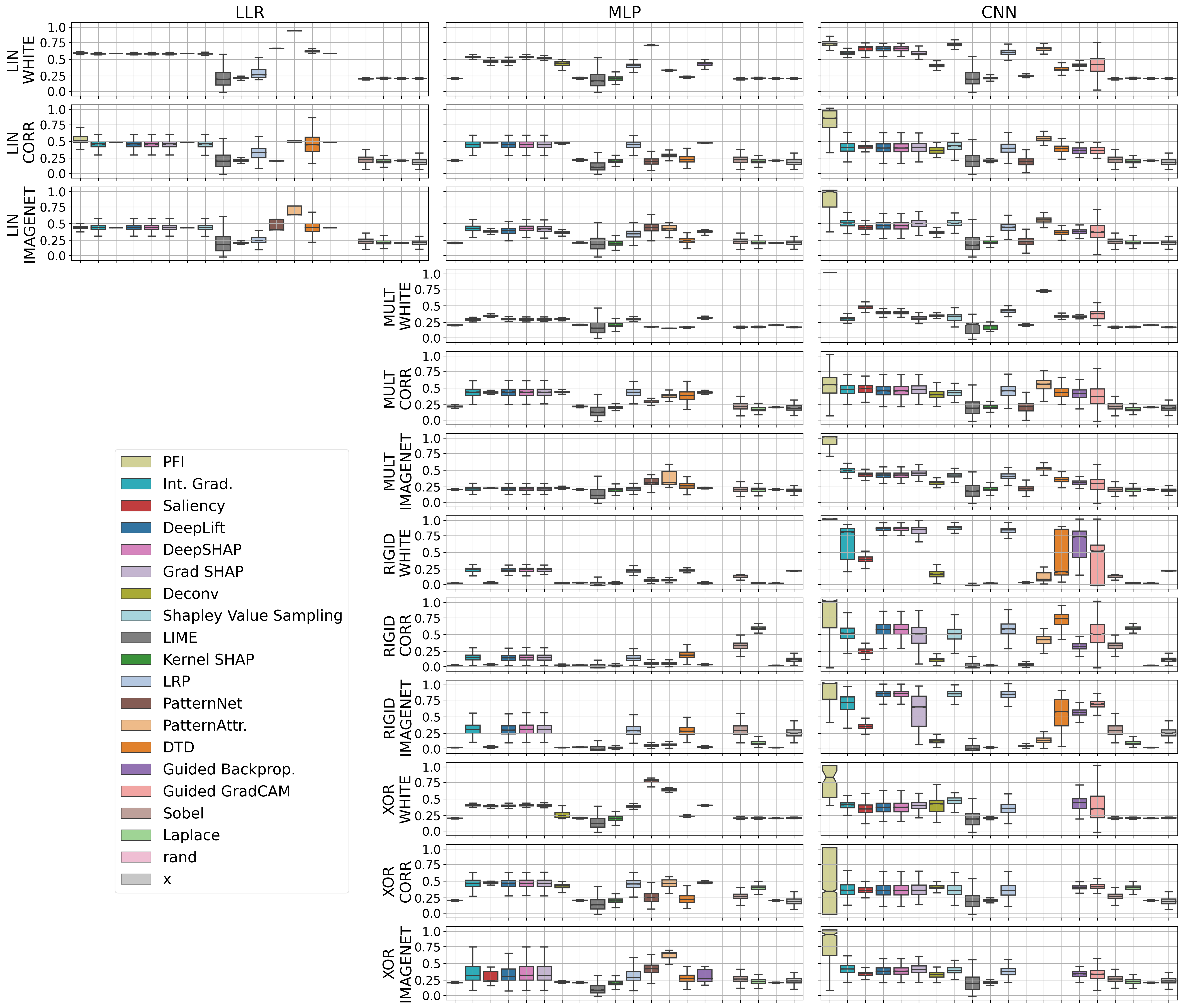}
    \caption{$\mathrm{IMA}$ metric results for every XAI method tested, separated by model architecture and depicted as boxplots of median and quartile performance scores. Guided GradCAM is only implemented for CNN architectures, and Keras models required for PatternNet, PatternAttribution, and Deep Taylor Decompostion (DTD) struggled to converge for the XOR scenarios as stated above, so these are excluded from the corresponding sub-plots. For the most part, results are relatively consistent with the above $\mathrm{EMD}$ results of Figure \ref{fig:emd-full-results}. Some methods see a drop in explanation performance as model complexity increases, from the Linear Logistic Regression (LLR) model to a Convolutional Neural Network (CNN). In the RIGID CORR case, the model-ignorant Laplace filter outright performs the best for explanations of MLP decisions and nearly so for the CNN. SHAP variants DeepSHAP, GradSHAP, and Shapley Value Sampling perform very similarly to one another in most cases across all model types, despite being formulated to target particular problems. One noticeable difference between the $\mathrm{EMD}$ results of Figure \ref{fig:emd-full-results} and the results shown here is that PatternAttribution performs outright best for LIN WHITE under the LLR and MLP, and XOR WHITE under the MLP. In contrast, PFI performs strongly for many scenarios under the CNN, but poorly under the MLP. No XAI method performs outright the best across all scenarios.}
    \label{fig:frac-full-results}
\end{figure}

\begin{figure}[!ht]
    \centering
    \includegraphics[width=\textwidth]{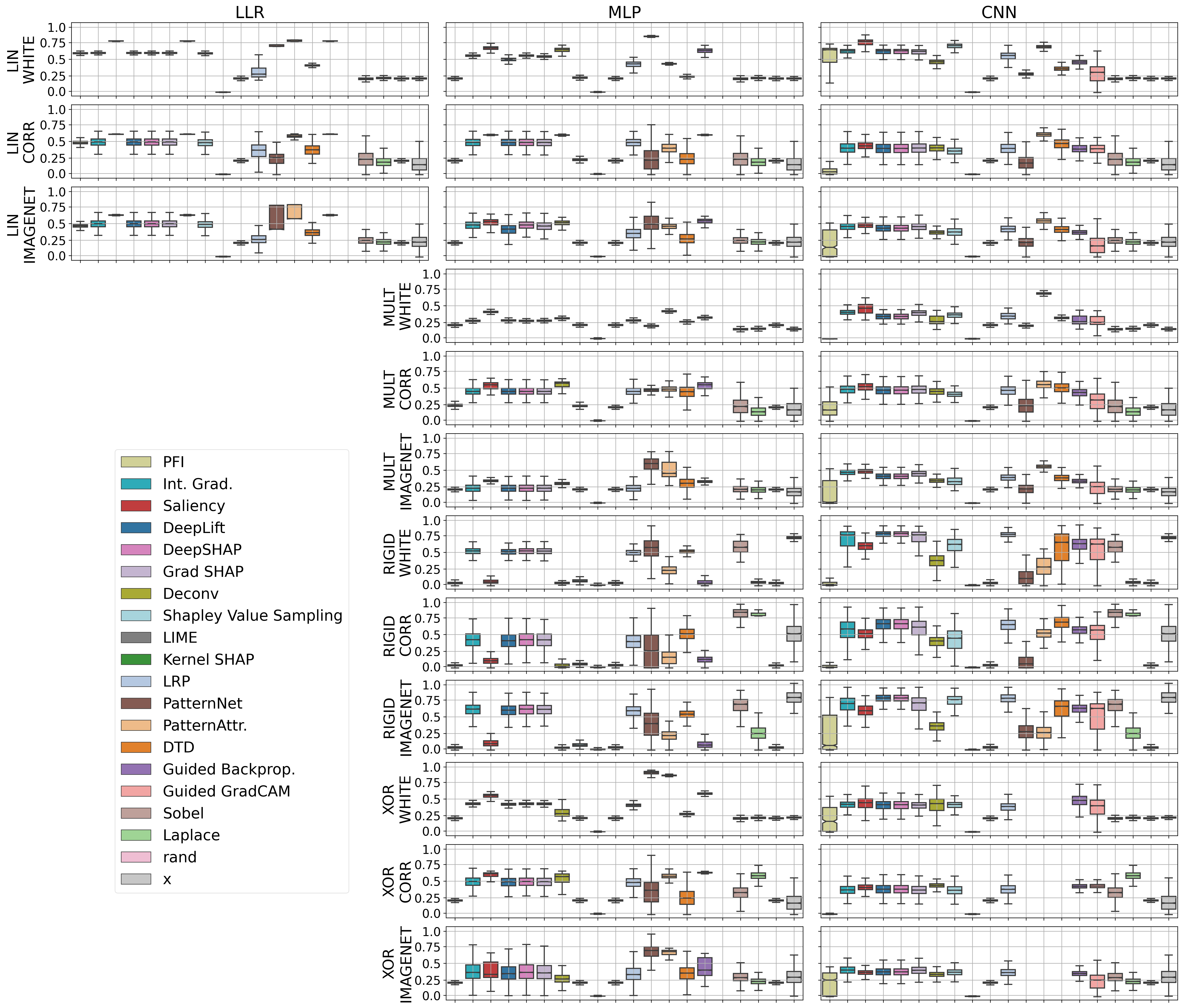}
    \caption{Precision score for every XAI method tested, separated by model architecture and depicted as boxplots of median and quartile performance scores. Most methods outperform the baseline methods for most model-scenario parameterization pairs. The `x' method, using input data as reference point of explanation, performs better for scenarios with higher signal-to-noise ratio (SNR), as the tetromino patterns will, on average, be more salient in the data there, thus present higher precision on average. Namely, the RIGID WHITE and IMAGENET scenarios generally require a higher SNR to be appropriately modeled. PatternNet and PatternAttribution, designed to nullify the influence of suppressor variables, generally perform well in the LIN and XOR WHITE cases, similar to the results shown by \citet{wilmingScrutinizingXAIUsing2022}, however these methods struggle in various other non-linear problem scenarios. LIME struggles across all scenarios, but performs better in the results shown in supplementary materials Section \ref{sup:8by8}, with the smaller $8 \times 8$-px image benchmark. Similarly to the results of \ref{fig:emd-full-results}, no XAI method performs outright the best across all scenarios. }
    \label{fig:precision-full-results}
\end{figure}

In Figures \ref{fig:emd-full-results}, \ref{fig:frac-full-results}, and \ref{fig:precision-full-results} we can see the full quantitative results for the $\mathrm{EMD}$, $\mathrm{IMA}$, and Precision metrics respectively, across all XAI methods and baselines. We can also see results for the PatternNet, PatternAttribution, and Deep Taylor Decomposition (DTD) methods, which are part of the Keras-based iNNvestigate framework \citep{alberINNvestigateNeuralNetworks2018}.

\subsection{8x8 Benchmarks}\label{sup:8by8}
\begin{figure}[!ht]
    \centering
    \begin{subfigure}{\textwidth}
        \centering
        \includegraphics[width=0.9\linewidth]{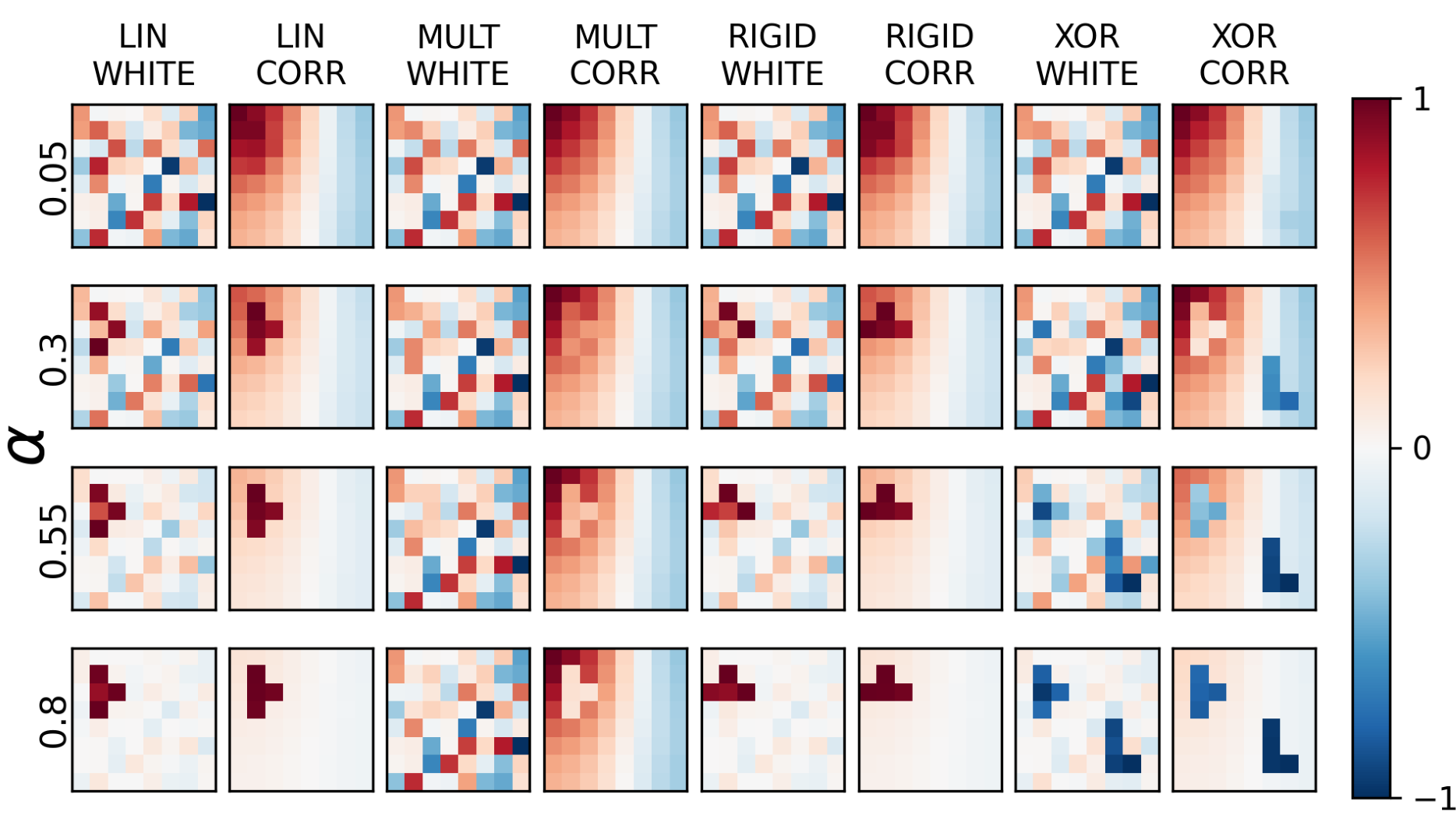}
        \caption{One generated sample of Class \#0 (where y=0) for four different SNRs $\alpha$.}
    \label{fig:snr_plot_subfig8x8}
    \end{subfigure} 
    \begin{subfigure}{\textwidth}
        \centering
        \includegraphics[width=0.88\linewidth]{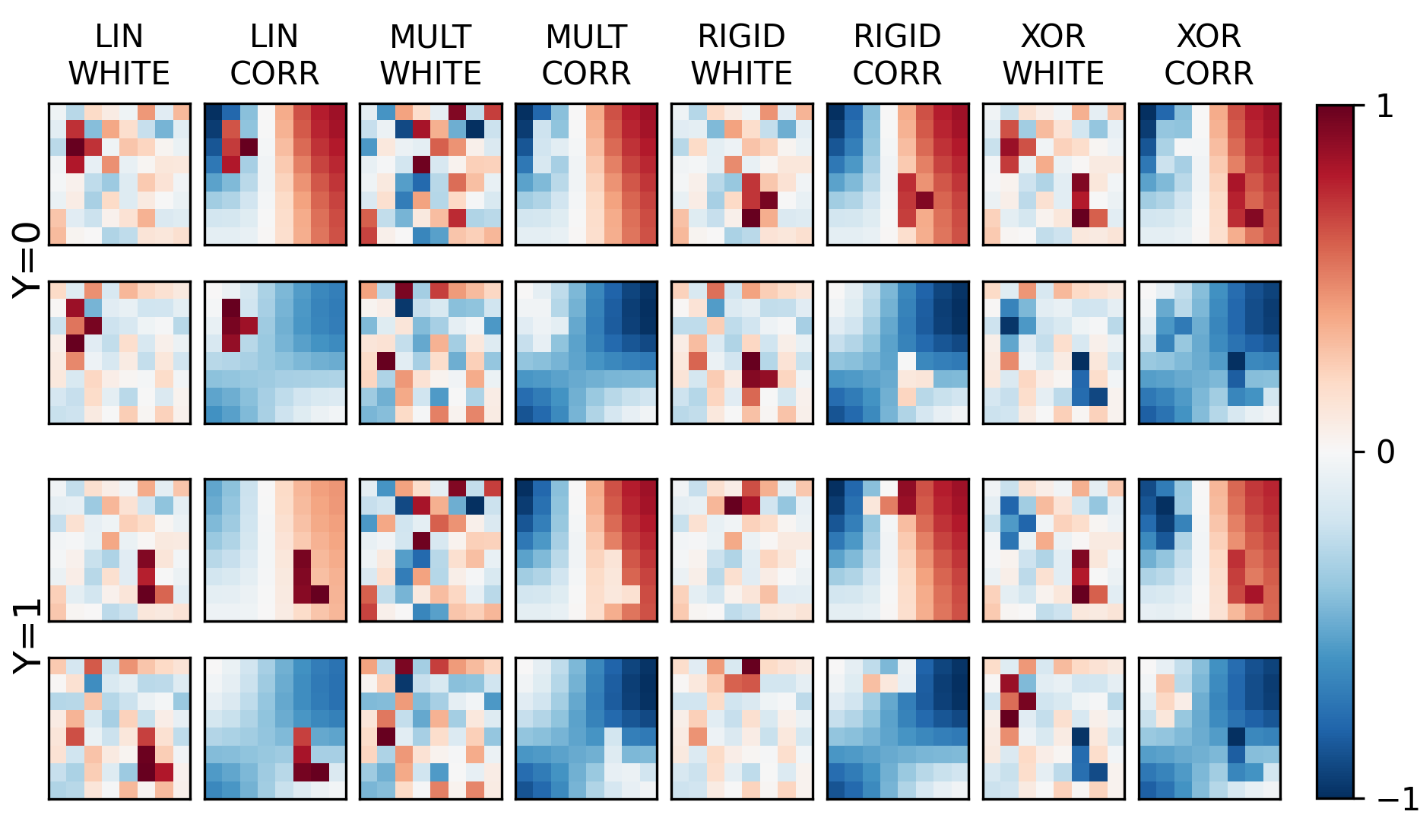}
        \caption{Two generated samples of each class per scenario.}
        \label{fig:data_plot_corrected8x8}
    \end{subfigure}
    \caption{Examples of generated $8 \times 8$-px data samples for each scenario, showing how an example for each scenario varies across four signal-to-noise ratios (SNRs) $\alpha$ (top).}
    \label{fig:snr_plot8x8}
    \vskip -0.1in
\end{figure}

The benchmark was originally designed around $8 \times 8$-px tetromino images, scaled up to $64 \times 64$-px with the inclusion of the ImageNet data as a third background type. This was done to improve the robustness and real-world applicability of the datasets and benchmarks present in this work. The original results for the  $8 \times 8$-px data with $1$-px thick tetrominoes can be seen in this section. 
Figure \ref{fig:snr_plot8x8} shows example data for both classes and also across a range of four $\alpha$ values. 
For CORR backgrounds, we set $\sigma_{\text{smooth}}=3.0$ for the smoothing filter, and no pattern smoothing was incorporated.
Here, each scenario was constructed with sample size $N=10,000$ and with an $80/10/10$ train/val/test split, with 25 datasets per scenario being used for analyses.

The Linear Logistic Regression (LLR) model in these experiments was the same single-layer neural network with two output neurons and a softmax activation function. 
The Multi-Layer Perceptron (MLP) similarly has four fully-connected layers and Rectified Linear Unit (ReLU) activations, and each of the fully-connected hidden layers halves the input size, i.e. [64, 32, 16, 8]. 
The two-neuron output layer was once again softmax-activated. 
Finally, the Convolutional Neural Network (CNN) was defined as four blocks of ReLU-activated convolutional layers followed by a max-pooling operation, with a softmax-activated two-neuron output layer. The convolutional layers are specified with four filters, a kernel size of two, a stride of one, and padding such that the input and output shapes match. 
This padding technique was used to improve pixel utilization across each convolution, as well as to mitigate shrinking outputs of the already relatively small images, by adding extra filler pixels (set to values of zero) around the edge of each image. 
The max-pooling layers are defined with a kernel size of two and a stride of two. 
As with the CNN architecture of the main text, some popular CNN architecture features (such as batch normalization) are unavailable here due to lack of implementation support by some XAI methods. 

\begin{figure}[!ht]
    \centering
    \includegraphics[width=\linewidth]{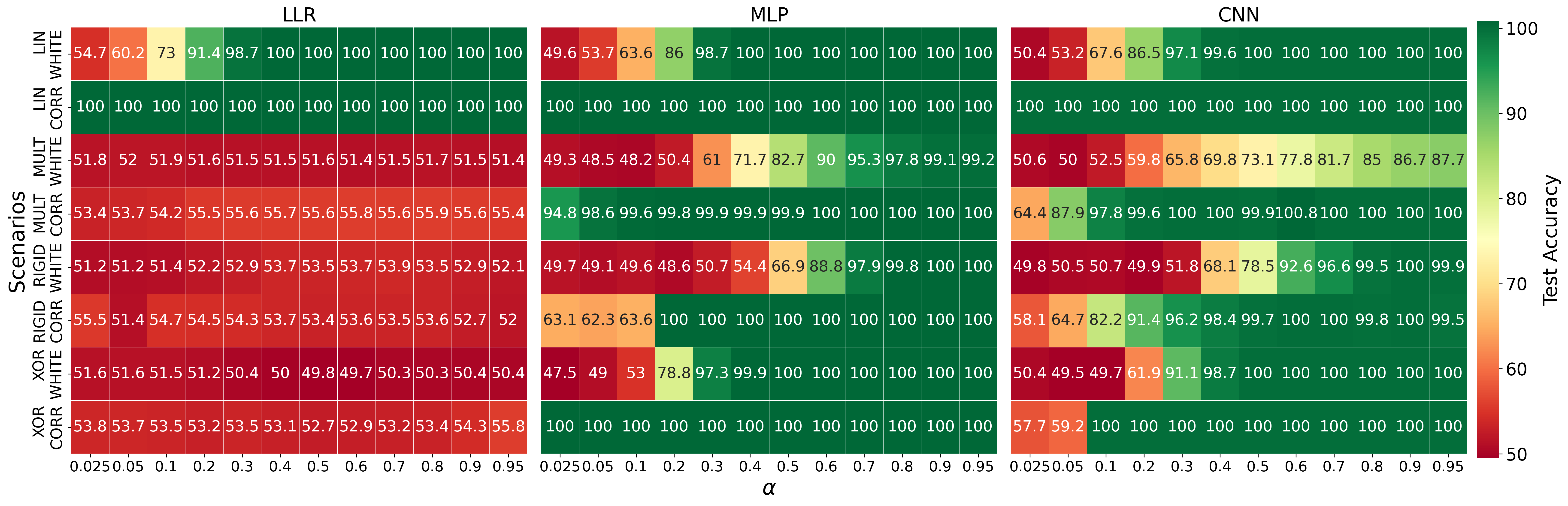}
    \caption{Average test accuracy over 10 model trainings for each problem scenario and model architecture of the $8 \times 8$-px setting, for a fixed range of signal-to-noise ratios (SNRs). As expected, the Linear Logistic Regression (LLR) model cannot perform above chance level for non-linear scenarios. The Convolutional Neural Network (CNN) would be expected to outperform the Multi-Layer Perceptron (MLP) for the RIGID (translations and rotations of tetrominoes) scenarios due to the invariance under these properties for this architecture. However, performance is comparable, with the MLP obtaining an average test accuracy above the 80\% threshold at a lower SNR than the CNN. This may be partially due to the compromise in the architecture of the CNN, where we were not able to use Batch Normalization due to incompatibility with some XAI frameworks and methods.}
    \label{fig:training-results8x8}
\end{figure}

Figure \ref{fig:training-results8x8} shows the training results across ten $\alpha$ values along with Table \ref{tab:chosen-snrs8x8} which shows the chosen $\alpha$ values used for analysis. 
Each network was trained over 500 epochs using the Adam optimizer without regularization, with a learning rate of $0.004$ for the LIN, MULT, and XOR scenarios, and $0.0004$ for the RIGID scenario. 

\begin{figure}[!ht]
    \centering
    \includegraphics[width=\textwidth]{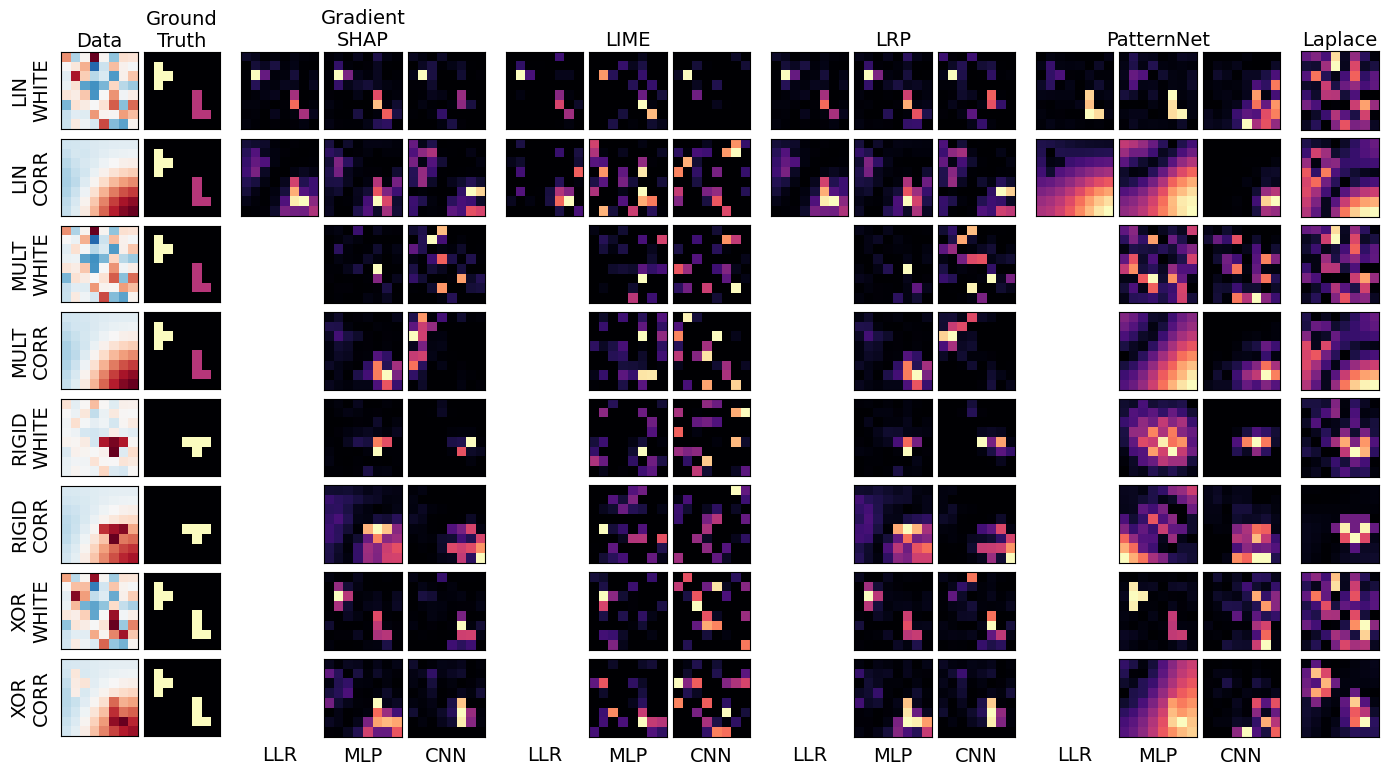}
    \caption{Absolute-valued importance maps obtained for a random correctly-predicted $8 \times 8$-px data sample, for selected XAI methods and baselines. Recovery of the ground truth pattern across all scenarios is best shown by XAI methods applied to a Linear Logistic Regression (LLR) model.
    % The Multi-Layer Perceptron (MLP) tends to focus on noise in the case of ImageNet backgrounds, and LIME often fails to produce sensical explanations across all model architectures.
    % Note that we have also highlighted $\mathbf{a}^{\text{L}}$ in the ground truth for the LIN and MULT examples. We consider the opposing-class tetromino pattern $\mathbf{a}^{\text{L}}$ as a member of $\mathcal{F}^+$, the ground truth set of important features, because the absence of these features can be a valid reason for an XAI to highlight the respective region of the image as important.
    }
    \label{fig:qualitative-results8x8}
    \vskip -0.2in
\end{figure}

\begin{figure}[!ht]
    \centering
    \includegraphics[width=0.7\textwidth]{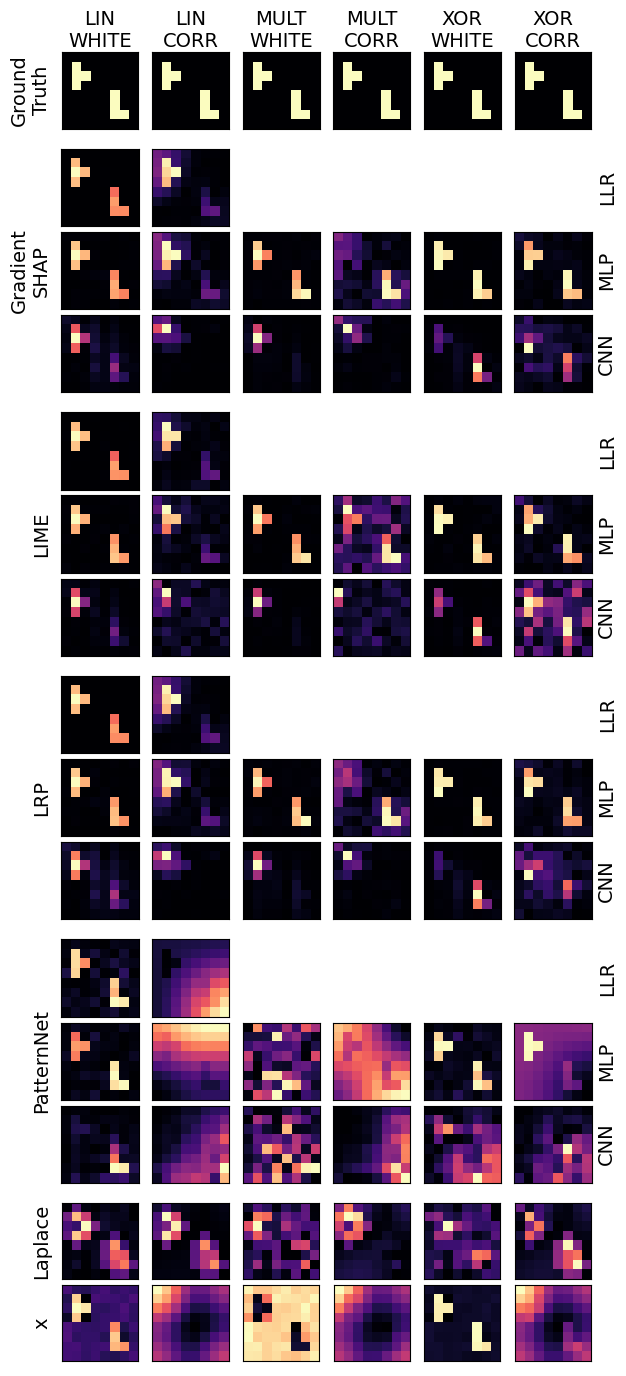}
    \caption{Absolute-valued global importance maps calculated as the mean importance value over all correctly predicted $8 \times 8$-px scenario samples, for selected XAI methods and baselines. RIGID scenarios involving translations and rotations of the tetromino signal pattern are not included as they have no fixed ground truth position. CORR scenarios with correlated background can be seen to produce noisier global importance maps, suggesting that this setting induces suppressor variables in the background, which are difficult for XAI methods to distinguish from the true signal pattern.}
    \label{fig:qualitative-results-global8x8}
\end{figure}

\begin{figure}[!ht]
    \centering
    \includegraphics[width=\textwidth]{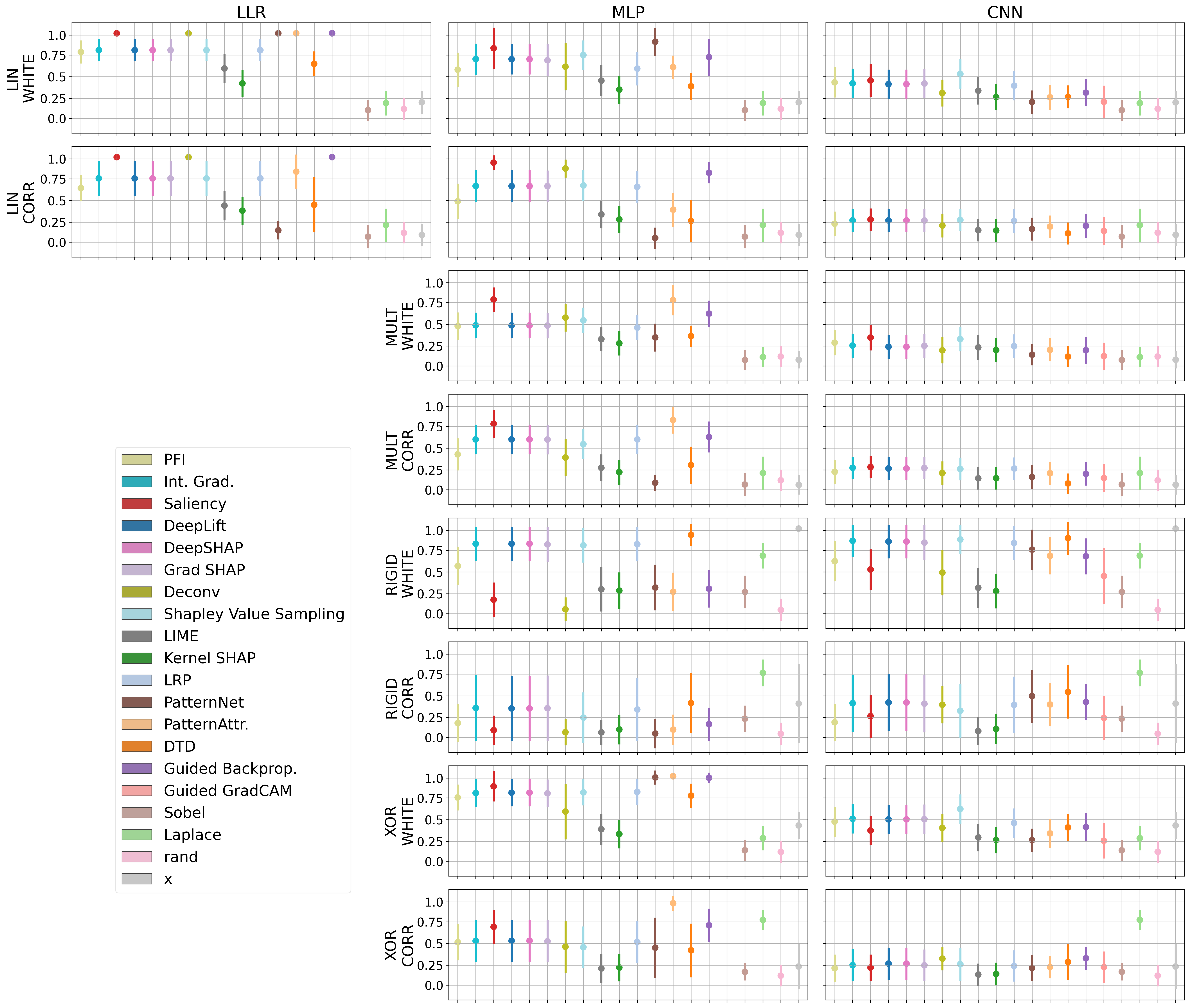}
    \caption{Precision score for every XAI method tested in the $8 \times 8$-px setting, separated by model architecture and depicted as mean and standard deviation performance scores. Most methods outperform the baseline methods for most model-scenario parameterization pairs. The `x' method, using input data as reference point of explanation, performs better for scenarios with higher signal-to-noise ratio (SNR), as the tetromino patterns will, on average, be more salient in the data there, thus present higher precision on average. Namely, the RIGID and WHITE scenarios generally require a higher SNR to be appropriately modeled. Outside of this, performance for XAI methods for the Convolutional Neural Network (CNN) is comparable to baseline methods.}
    \label{fig:precision-full-results8x8}
\end{figure}

Figures \ref{fig:qualitative-results8x8} and \ref{fig:qualitative-results-global8x8} show qualitative results for local and global explanations respectively, and Figures \ref{fig:emd-full-results8x8} and \ref{fig:precision-full-results8x8} show quantitative results for the $\mathrm{EMD}$ and Precision metrics respectively.

\begin{table}[!ht]
    \caption{Results of the model training process for each classification setting, model architecture, and background type in the $8 \times 8$-px setting. These results are depicted as chosen Signal-to-noise ratios (SNRs), parameterized by $\alpha$, as well as the average test accuracy (ACC, \%). }
\label{tab:chosen-snrs8x8}
\vskip 0.15in
\begin{small}
\begin{sc}
\begin{tabular}{lccccr}
\toprule
& & \multicolumn{2}{c}{WHITE} & \multicolumn{2}{c}{CORR} \\
& & $\alpha$ & ACC & $\alpha$ & ACC \\
\midrule
        &       LLR     & $0.1800$ & $88.9$ & $0.0125$ & $99.9 $ \\
LIN     &       MLP     & $0.1800$ & $87.9 $ & $0.0125$ & $99.9$ \\
        &       CNN     & $0.1800$ & $83.0 $ & $0.0125$ & $86.4$ \\
\rule{0pt}{2.5ex}MULT    &    MLP        & $0.7000$ & $93.6$  & $0.1000$ & $99.4$ \\
        &    CNN        & $0.7000$ & $83.1$  & $0.1000$ & $90.6 $ \\
\rule{0pt}{2.5ex}RIGID   &    MLP        & $0.6500$ & $91.9$ & $0.2000$ & $99.9$ \\
        &    CNN        & $0.6500$ & $93.7$ & $0.2000$ & $88.8$ \\
\rule{0pt}{2.5ex}XOR     &    MLP        & $0.3500$ & $99.5$ & $0.1500$ & $100.0$ \\
        &    CNN        & $0.3500$ & $95.2$ & $0.1500$ & $99.5$ \\
\bottomrule
\end{tabular}
\end{sc}
\end{small}
% \vskip -0.25in
\end{table}

\begin{figure*}[!ht]
    \centering
    \includegraphics[width=\textwidth]{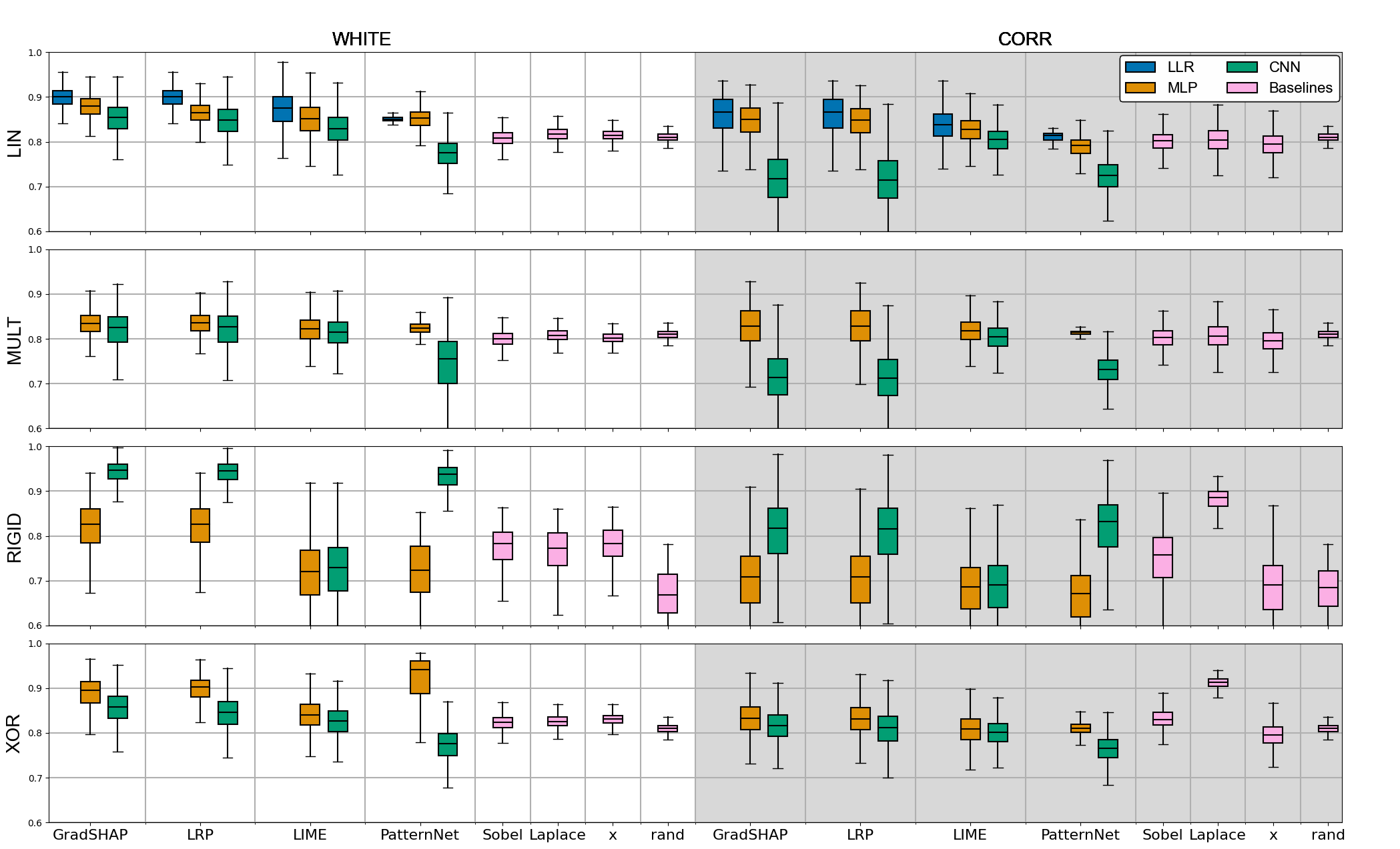}
    \caption{Quantitative explanation performance of individual sample-based feature importance maps produced by various XAI approaches and baseline methods on correctly-predicted $8 \times 8$-px scenario test samples, as per the $\mathrm{EMD}$ metric. 
    Depicted are boxplots of median explanation performance, with upper and lower quartiles as well as outliers shown. 
    The white area (left) shows results for white background noise (WHITE), whereas the gray shaded area (right) shows results for the correlated background noise (CORR) scenarios. 
    Explanation performance decreases as model complexity (from LLR to MLP to CNN) increases, with the exception of the RIGID scenarios, where the CNN is better suited to the non-static ground truth patterns present. Unlike results seen for linear data \citep{wilmingScrutinizingXAIUsing2022}, PatternNet and PatternAttribution do not outright outperform other XAI methods for most configurations. }
    \label{fig:emd-results8x8}
    % \vskip -0.1in
\end{figure*}

\begin{figure}[!ht]
    \centering
    \includegraphics[width=\textwidth]{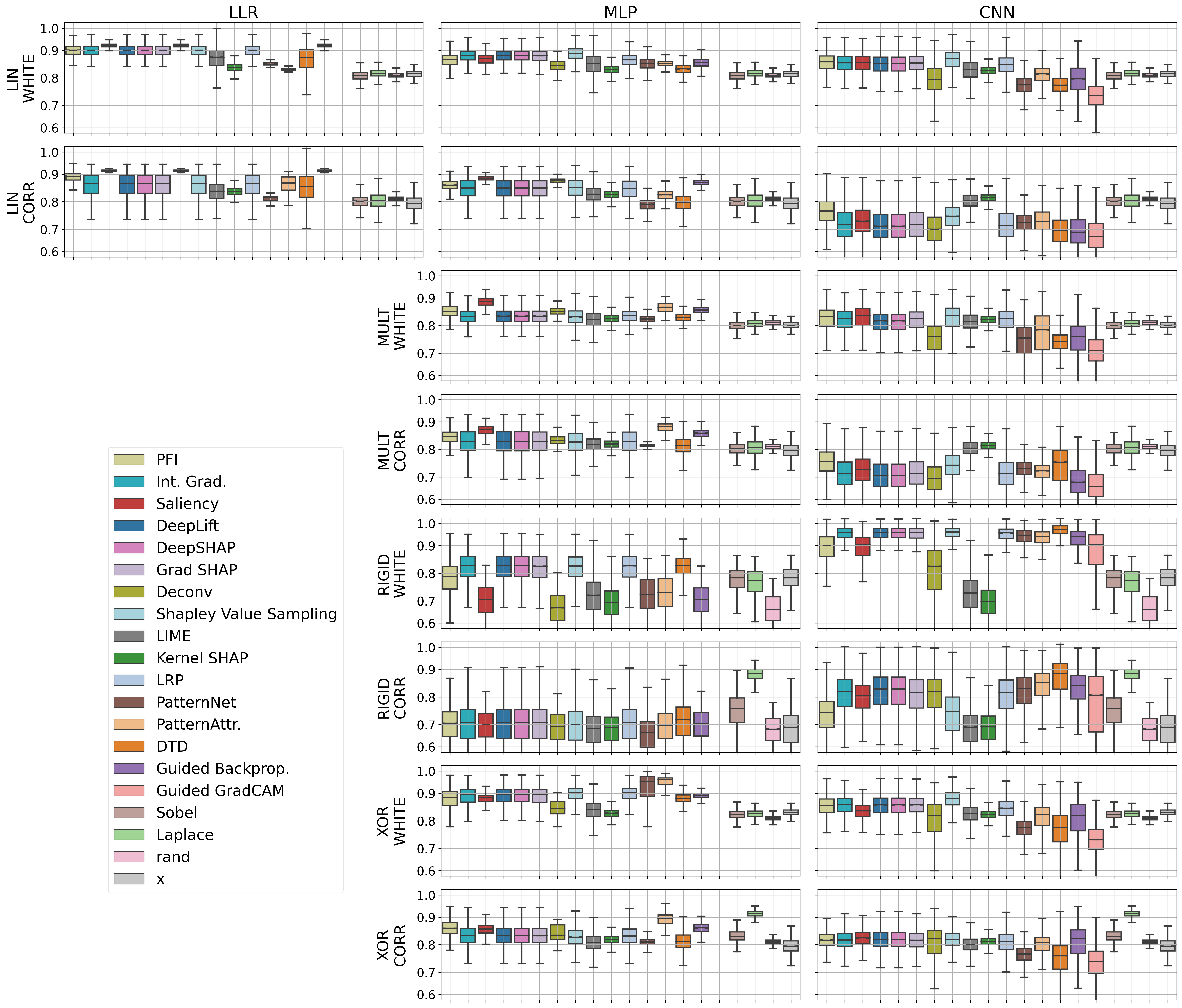}
    \caption{$\mathrm{EMD}$ metric based on the Earth Mover's Distance (EMD) for every XAI method tested in the $8 \times 8$-px setting, separated by model architecture and depicted as boxplots of median and quartile performance scores. Consistent with the results of Figure \ref{fig:emd-results8x8}, explanation performance tends to decrease as model complexity increases, from the Linear Logistic Regression (LLR) model to a Convolutional Neural Network (CNN). An exception is seen for RIGID scenarios where most XAI methods outperform the Multi-Layer Perceptron (MLP) equivalent. In this case, the model-ignorant Laplace filter performs the best across both architectures.}
    \label{fig:emd-full-results8x8}
\end{figure}

\clearpage
\newpage

\end{appendices}

\end{document}